\documentclass[twocolumn]{article}
\usepackage{graphicx} 
\usepackage[margin=2cm]{geometry}
\usepackage{amsmath}
\usepackage{mathtools}
\usepackage{tikz}
\usepackage{subcaption}
\usepackage{enumitem}
\usepackage[T1]{fontenc}
\usepackage{authblk}
\usepackage[hidelinks]{hyperref}
\usepackage{titlesec}
\usepackage{textgreek}

\titleformat{\section}
  {\normalfont\Large\bfseries\raggedright
   \hyphenpenalty=10000\exhyphenpenalty=10000}
  {\thesection}
  {1em}
  {}

\titleformat{\subsection}
  {\normalfont\large\bfseries\raggedright
   \hyphenpenalty=10000\exhyphenpenalty=10000}
  {\thesubsection}
  {1em}
  {}

\titleformat{\subsubsection}
  {\normalfont\normalsize\bfseries\raggedright
   \hyphenpenalty=10000\exhyphenpenalty=10000}
  {\thesubsubsection}
  {1em}
  {}

\definecolor{flatorange}{RGB}{230, 126, 34}
\definecolor{flatgreen}{RGB}{39, 174, 96}
\definecolor{flatpurple}{RGB}{142, 68, 173}
\definecolor{flatgray}{RGB}{127, 140, 141}
\definecolor{flatred}{RGB}{192, 57, 43}
\definecolor{flatyellow}{RGB}{241, 196, 15}
\definecolor{flatblue}{RGB}{41, 128, 185}
\definecolor{flatblue2}{RGB}{52, 152, 219}
\definecolor{flatlaci}{RGB}{44, 62, 80}
\definecolor{flatlaci2}{RGB}{52, 73, 94}

\title{Quantum-Inspired Evolutionary Algorithm with Graph-Based Qudit Representation and Heuristic Initialization}

\author[1]{Umut Çalıkyılmaz}
\author[1]{Nitin Nayak}
\author[2]{Sven Groppe}

\affil[1]{University of Lübeck, Lübeck, Germany}
\affil[2]{TU Bergakademie Freiberg, Freiberg, Germany}

\affil[ ]{\small
\texttt{umut.calikyilmaz@uni-luebeck.de},
\texttt{nitin.nayak@uni-luebeck.de}\\

\texttt{sven.groppe@informatik.tu-freiberg.de}}

\date{November 2026}

\begin{document}

\maketitle

\begin{abstract}

Recent variants of the quantum-inspired evolutionary algorithm (QEA) have adopted qudit-based representations for integer solution spaces and permutation-preserving measurement techniques for ordering problems. We build on such studies to propose two variants: MQEA-G, which incorporates a graph-based qudit model, and MQEA-GH, which extends this model with heuristic initialization. We implement both variants along with binary QEA (BQEA) and baseline multi-level QEA (MQEA) within a common algorithmic framework, offering a fine-grained configuration space with configurable model-best, iteration-best, and global-best rotations. Experiments on synthetic and benchmark traveling salesperson problem instances show that the best-performing configurations of both proposed variants outperform those of BQEA and MQEA across all evaluated instances and time limits, with MQEA-GH further outperforming MQEA-G. Compared with the tested configurations of ant colony optimization, genetic algorithms, and simulated annealing, MQEA-GH configurations perform particularly well for synthetic instances while also being competitive for benchmark instances. Configuration analysis highlights the importance of heuristic initialization and qudit-amplitude constraining, as well as the context dependence of rotation settings.
\end{abstract}

\section{Introduction}
\label{sec:introduction}

The traveling salesperson problem (TSP) is a well-known combinatorial optimization problem with the objective of finding the shortest Hamiltonian cycle on a weighted, undirected, complete graph \cite{gavish1978travelling}. It is named after an imaginary salesperson who is supposed to visit $n$ cities in a closed loop while following the shortest possible path according to the given distances between the cities. As the TSP is NP-hard \cite{karp1972reducibility}, numerous metaheuristic approaches have been applied to it, including ant colony optimization (ACO) \cite{dorigo1997ant}, the genetic algorithm (GA) \cite{grefenstette1985genetic}, and simulated annealing (SA) \cite{aarts1988quantitative}.

The quantum-inspired evolutionary algorithm (QEA) is a metaheuristic method that uses quantum-inspired probabilistic representations of candidate solutions, aiming to balance exploration and exploitation \cite{han2000genetic,han2002quantum}. The original algorithm and most of its variants use qubit representations \cite{shu2007quantum,li2005independent}. We refer to such methods as binary QEA (BQEA). Some more recent variants rely on qudits (multi-level quantum systems) instead of two-level qubits for a more expressive encoding of candidate solutions \cite{tkachuk2018quantum,saitoh2025quantum}, which we refer to as multi-level QEA (MQEA). Similarly, some studies proposed quantum-inspired methods specializing in permutation search spaces, where the model measurement operation is modified to produce only valid permutations \cite{da2017quantum,soloviev2021quantum}.

Building on permutation-preserving measurement, we propose two novel mechanisms for MQEA to solve ordering problems, such as the TSP. The first is a graph-based qudit representation, in which the outcome of each measurement determines the next qudit to be measured, instead of using a fixed order and measuring the qudits sequentially. Similarly, when updating a graph-based model toward a guiding solution, each integer value indicates the next qudit to be rotated. This structure enables the probabilistic representation to capture dependencies between consecutive choices, which can be particularly useful for problems with permutation solution spaces. In exchange, it requires only one additional qudit and does not require hard-to-simulate entangled states. The second is heuristic-based initialization, in which problem-specific information is used to determine the initial coefficients of the qudits. This mechanism introduces a configurable heuristic influence hyperparameter, which determines the impact of the heuristic information during initialization. We denote the variant incorporating both proposed mechanisms as MQEA-GH. We also introduce the intermediate variant using only the graph-based model, MQEA-G, to better distinguish the effectiveness of the two mechanisms.

We develop a common algorithmic framework for BQEA, MQEA, MQEA-G, and MQEA-GH, incorporating three types of rotations guided by model-best, iteration-best, and global-best solutions. The framework offers configurable rotation counts and base rotation angles for each rotation type, angle-calculation operators, and a hyperparameter for constraining qudit amplitudes. We integrate this implementation into the STILO framework \cite{ccalikyilmaz2026metaheuristic}, which is a metaheuristic optimization framework specifically built for time-limited optimization. The resulting framework, STILO-Q\footnote{\url{https://github.com/umutcalikyilmaz/STILO-Q}}, extends STILO with a wide range of BQEA, MQEA, MQEA-G and MQEA-GH configurations alongside its existing ACO, GA, and SA implementations. It also provides an environment for time-limited optimization and enables comparisons across different discrete problem instances and computational budgets.

Beyond evaluating solution quality under a single computational budget, we consider execution time as an additional dimension in algorithm evaluation, motivated by the requirements of real-time optimization. Real-time optimization is used in various domains such as multi-agent path-finding \cite{yukhnevich2026enhancing} and transaction scheduling \cite{ccalikyilmaz2025optima}, where execution times must satisfy strict limits. Iterative metaheuristics are particularly suitable for time-limited optimization, as their execution can be stopped after any number of iterations. For each tested algorithm family, we record the objective value obtained by the best-performing configuration under each tested time limit, ranging from 50 ms to 1000 ms. We use synthetically generated instances for an analysis based on instance characteristics, as well as benchmark instances to produce reproducible comparisons.

The experimental results demonstrate that both proposed variants outperform BQEA and MQEA across all tested settings, with MQEA-GH further outperforming MQEA-G. Furthermore, MQEA-GH was competitive with the established metaheuristics ACO, GA, and SA, achieving better solution quality in several settings. Configuration analysis reveals a general preference for high heuristic influence values and positive constraints on qudit amplitudes, while showing the preferred settings for each rotation type depend on instance characteristics and computational budgets.

The remainder of the paper is organized as follows. Section~\ref{sec:qea} introduces the qubit representation, quantum-inspired evolutionary algorithm, and existing variants. Section~\ref{sec:mqea} introduces the qudit-based variant and the common algorithm framework used in this study while also explaining the proposed novel features. Section~\ref{sec:experimental} presents the results of the experimental evaluation comparing the relative performances of the tested methods and showing the impact of the proposed mechanisms. Section~\ref{sec:conclusion} summarizes the main contributions and outlines directions for future work.
 
\section{Quantum-Inspired Evolutionary Algorithm}
\label{sec:qea}

We first introduce the qubit-based representation and describe the algorithm structure of the original BQEA \cite{han2000genetic, han2002quantum}. We then review the existing QEA variants, including those based on qudits.

\subsection{Qubit Representation}
\label{ssec:qubit}

\begin{figure}
    \begin{subfigure}[b]{0.225\textwidth}    
    
    \centering
	\begin{tikzpicture}
        \draw[line width=0.6] (0,0) circle (1.4);
        \draw[->, line width=0.8] (0,0)--(1.9,0);
        \node at (1.9,0.2) {\footnotesize $y$};
        \draw[->, line width=0.8] (0,0)--(0,1.9);
        \node at (0.2,1.9) {\footnotesize $z$};
        \draw[->, line width=0.8] (0,0)--(-0.6,-0.6);
        \node at (-0.4,-0.7) {\footnotesize $x$};
        
        \draw[dashed] (-1.4,0) .. controls (-1,0.5) and (1,0.5) .. (1.4,0);
        \draw[dashed] (-1.4,0) .. controls (-1,-0.5) and (1,-0.5) .. (1.4,0);
        \draw (0,0) -- (0.4,0.95);
        \draw[densely dotted, line width=0.6] (0.4,-0.2) -- (0.4,0.95);
        \draw[densely dotted, line width=0.6] (0.4,-0.2) -- (0,0);

        \filldraw [black] (0,1.4) circle (1.5pt);
        \node at (-0.23,1.57) {\footnotesize $\left|0\right\rangle$};
        \filldraw [black] (0,-1.4) circle (1.5pt);
        \node at (0,-1.63) {\footnotesize $\left|1\right\rangle$};
        \filldraw [black] (0.4,0.95) circle (1pt);
        \node at (0.65,0.95) {\footnotesize $\left|\psi\right\rangle$};

        \draw (0,0.5) arc (90:68:0.5);
        \node at (0.12,0.6) {\tiny $\upsilon$};

        \draw (-0.1,-0.1) arc (-110:-65:0.33);
        \node at (0.03,-0.23) {\tiny $\phi$};
    \end{tikzpicture}
    \caption{Bloch Sphere Representation}
    \label{fig:Bloch}
    \end{subfigure}    
    \hfill    
    \begin{subfigure}[b]{0.225\textwidth}
        \centering
        \begin{tikzpicture}
        \draw[->, line width=1] (0,0)--(2.5,0);
        \draw[->, line width=1] (0,0)--(0,2.5);
        \draw[->, line width=0.8] (0,0)--(1.5,2);
        \draw (0,0.5) arc (90:52:0.5);

        \node at (0.3, 2.5) {\footnotesize $\left|0\right\rangle$};
        \node at (2.75,0.1) {\footnotesize $\left|1\right\rangle$};
        \node at (1.75,2.2) {\footnotesize $\left|\psi\right\rangle$};
        \node at (0.2,0.65) {\footnotesize $\theta$};
                    
    \end{tikzpicture} 
    \caption{Real-Amplitude Representation}
    \label{fig:alternative}
    \end{subfigure}
    \caption{Two alternative qubit representations}
\end{figure}
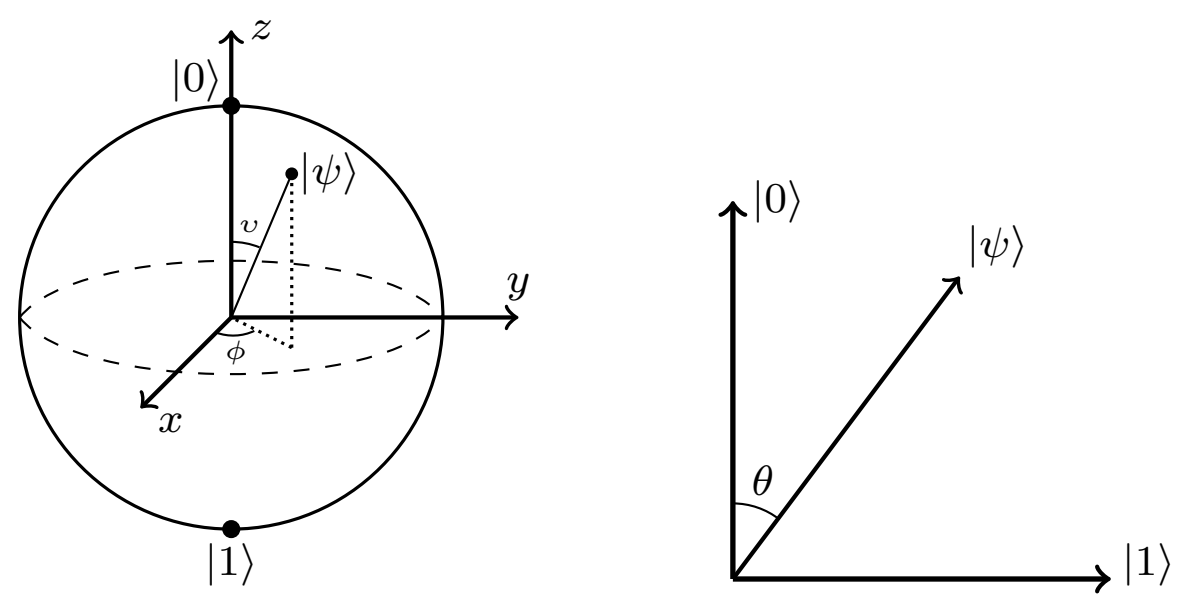

A qubit is a two-level quantum system and the smallest unit of quantum information. Similar to its classical analogue, it can give one of the two possible outcomes when observed. However, unlike a classical bit, a qubit can be in a \textit{superposition} state, shown as:

\begin{equation}
    \left|\psi \right\rangle = \alpha  \left|0 \right\rangle + \beta \left|1 \right\rangle \equiv \left[\alpha,\beta \right]^T
        \label{eq:qubitstate}
\end{equation}
where $\alpha$ and $\beta$ are complex values called \textit{probability amplitudes}, which satisfy the normalization condition $|\alpha|^2+|\beta|^2=1$. Measuring such a qubit yields either $0$ or $1$, with probabilities $|\alpha|^2$ and $|\beta|^2$, respectively. The general qubit state has two degrees of freedom, since the normalization condition and the physical irrelevance of the global phase reduce the four real parameters of the two complex amplitudes by two. Therefore, a qubit state can be represented by a point on a Bloch sphere, as shown in Figure~\ref{fig:Bloch}.

A system of multiple qubits can also be in a superposition of computational basis states, such as the two-qubit state given below.

\begin{equation}
    \left|\psi \right\rangle = c_{00}  \left|00 \right\rangle + c_{01} \left|01 \right\rangle + c_{10} \left|10 \right\rangle + c_{11} \left|11 \right\rangle
        \label{eq:twoqubitstate}
\end{equation}
In general, representing an arbitrary state of $n$ qubits requires $2^n$ complex amplitudes. A system with multiple qubits can be in an \textit{entangled} state, which cannot be represented as the product of the states of individual qubits. In contrast, a \textit{product state} does not have entanglement and can be represented only by the probability amplitudes of each qubit. Thus, such a system can be represented using only $2n$ complex amplitudes.

To simplify simulation, QEA uses a product state of multiple qubits as the probabilistic model. Moreover, the amplitudes of each qubit are restricted to real values instead of complex ones, which reduces the degrees of freedom of each qubit to one. With this simplification, each qubit can be represented by a vector in the two-dimensional plane, as shown in Figure~\ref{fig:alternative}, and described by a single angle $\theta$:

\begin{equation}
    \left|\psi\right\rangle = \cos\theta \left|0\right\rangle + \sin\theta\left|1\right\rangle
\end{equation}

In QEA, multiple probabilistic models are maintained, each of which consists of a vector of simplified qubits. In each generation, the qubits in each model are \textit{measured} to produce binary strings, each representing a candidate solution. During the measurement of a qubit, a pseudo-random number $p$ is generated uniformly from $[0,1]$. The outcome is $0$ if $p<\cos^2\theta$, and $1$ otherwise. Since simulated qubits are used instead of real quantum systems, the state does not collapse after a measurement, and it can be measured multiple times.

Probabilistic models are updated by applying \textit{rotation gates} to the qubits. For a physical qubit, a rotation gate alters the state by rotating its Bloch vector around an axis on the Bloch sphere. For the simplified qubits used in QEA, the rotation operator $R(\Delta \theta)$ updates the angle according to $\theta\rightarrow\theta+\Delta\theta$.

\subsection{Algorithm Structure}

The algorithm described in \cite{han2002quantum} starts by initializing all the qubits of every model to an equal superposition state by setting each $\theta$ value to $\pi/4$. Therefore, the probabilities of the outcomes $0$ and $1$ are equal for each qubit. In each iteration, each model is measured to produce one or more binary candidate solutions, which are then evaluated. The probabilistic models are then updated using rotation operators guided by the obtained solutions. The rotation angles are determined using a lookup table based on the objective values and the corresponding bits of the guiding solution and the best solution obtained from the model so far.

At the end of each generation, the best solution for each model is updated, if necessary, as well as the global best solution. Afterwards, the algorithm may invoke a mechanism called \textit{migration}, which replaces the stored best solutions of selected models with either the best solution within a group of models (local migration) or the global best solution (global migration). The specific conditions to invoke local and global migration may depend on the particular application. Migration influences the direction of search by changing the solutions used to guide model updates.

\subsection{Existing Variants}

Since its introduction, many variants of QEA have been proposed for various domains. One major focus of research on QEA has been rotation-angle calculation. The original method employed a lookup table to determine the direction and magnitude of each rotation \cite{han2000genetic,han2002quantum}. Many later studies adopted this approach and proposed various lookup tables for different applications \cite{shu2007quantum,li2005independent}.
However, angle calculation based on lookup tables restricts the rotation angles to predefined values and often requires application-dependent update rules. Therefore, alternative approaches have been proposed. The authors of \cite{wang2013improvement} propose a method based on matrix multiplications to determine the rotation direction and an iteration-dependent calculation method for rotation-angle magnitude. Similarly, in \cite{vlachogiannis2009reactive, han2011quantum}, the magnitude of a rotation depends on the objective values of the best and measured solutions.

In addition to global and local migration, alternative mechanisms have been proposed for controlling the search direction in QEA. One such approach uses attractors that guide the search \cite{platel2007versatile}. In \cite{han2004quantum}, the authors of the original QEA papers proposed an operation called the $H_\epsilon$ gate to avoid premature convergence. The $H_\epsilon$ gate slightly rotates a qubit if its state is too close to either $\left|0\right\rangle$ or $\left|1\right\rangle$. Inspired by this mechanism, we introduce a configurable limit angle to constrain the qudit amplitudes and prevent excessive convergence toward individual outcomes.

Many hybrid methods combine operators from different metaheuristic approaches with QEA. One prominent idea is to integrate genetic algorithm operations into QEA, such as mutation and crossover, to increase the diversity of the candidate solutions \cite{mohammed2012quantum,li2005independent}. Others combine QEA with other metaheuristics, such as particle swarm optimization \cite{wang2007novel}, gravitational search \cite{ji2014application}, and cultural algorithms \cite{da2005cultural}. As QEA is often classified as a multimodel estimation of distribution algorithm \cite{platel2008quantum}, we focus on the probabilistic representations and model-update mechanisms rather than hybrid QEA approaches in this study.

For problems without a natural binary representation, binary encoding may cause redundancy, loss of locality, and poor scaling \cite{rothlauf2006representations}. In recent years, various optimization approaches have adopted qudit-based representations to produce integer solutions. These include qudit gradient descent and qudit local quantum annealing for graph coloring \cite{jansen2024qudit}, and a many-qudit formulation for the TSP \cite{vargas2021many}. Several QEA variants also use qudit-based representations for both continuous \cite{tkachuk2018quantum} and discrete problems \cite{saitoh2025quantum}. Similarly, other studies introduce specialized representations and measurement mechanisms for ordering problems, including the TSP \cite{da2017quantum,soloviev2021quantum}. Building on existing work on permutation-preserving quantum-inspired representations, we develop a qudit-based QEA approach and extend it with a graph-based model and heuristic initialization.

\section{Multi-Level Quantum-Inspired Evolutionary Algorithm}
\label{sec:mqea}

We implement the MQEA variants and BQEA in a common algorithm structure to investigate the impact of different probabilistic representations under the same conditions. Although the rotation operations described in this section are formulated for qudits, they are also applicable for BQEA, provided that qubit amplitudes are restricted to non-negative values. However, BQEA uses a different problem encoding, and its measurement operation produces binary strings rather than permutations. We point out the variations caused by these differences throughout this section.

We first introduce the qudit representation for MQEA variants and the two problem encodings used in this study. We then present the algorithm structure shared by the BQEA, baseline MQEA, MQEA-G, and MQEA-GH configurations. Finally, we describe the configuration components, including the proposed graph-based model and heuristic initialization. 

\subsection{Qudit Representation}
\label{ssec:qudit}

A qudit is a $q$-level quantum system that produces one of the $q$ possible outcomes when measured. The general state of a qudit can be shown as

\begin{equation}
    \left| \psi \right\rangle = \sum_{i\in\boldsymbol{O}} c_i \left| i \right\rangle,
\end{equation}
where all $c_i$ are complex probability amplitudes satisfying the normalization condition $\sum_{i\in\boldsymbol{O}} |c_i|^2=1$, $\boldsymbol{O}$ is the set of possible outcomes with $|\boldsymbol{O}|=q$, and $\left|i\right\rangle$ are orthonormal basis states. When the qudit is measured, it collapses into one of the basis states $\left|i\right\rangle$ with probability $|c_i|^2$, producing the outcome $i$.

Similar to BQEA, MQEA uses easy-to-simulate quantum-inspired states. For this purpose, only product states of multiple qudits are used, and their probability amplitudes are limited to real values. We further restrict the amplitudes to non-negative values in our representation. This eliminates the need for an additional mechanism to determine the rotation direction, such as a lookup table, while not restricting the representable probability distributions.

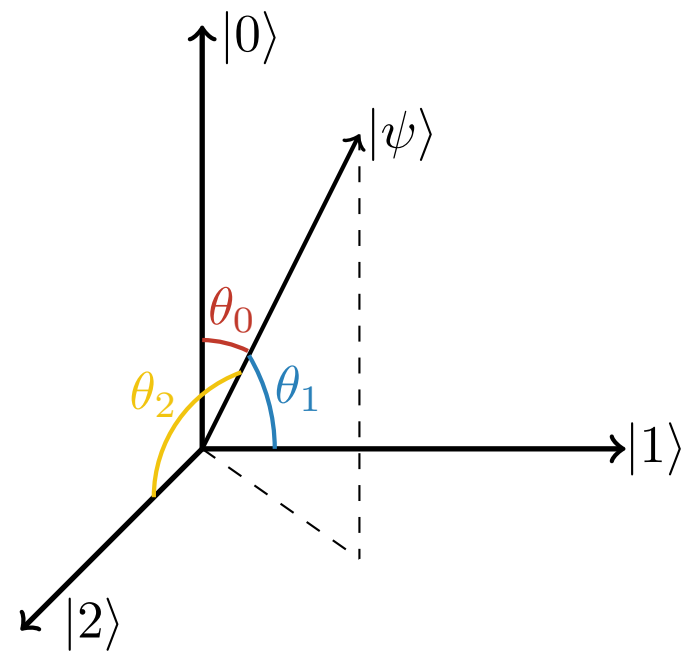
\begin{figure}
    \centering
	\begin{tikzpicture}[scale=0.8]
        \draw[->, line width=1] (0,0)--(3.5,0);
        \draw[->, line width=1] (0,0)--(0,3.5);
        \draw[->, line width=1] (0,0)--(-1.5,-1.5);
        \draw[->, line width=0.8] (0,0)--(1.3,2.6);
        \draw[dashed] (1.3,2.6)--(1.3, -0.91);
        \draw[dashed] (0,0)--(1.3, -0.91);
        \draw[color=flatred, line width=0.8] (0,0.9) arc (89:64:0.9);
        \draw[color=flatblue, line width=0.8] (0.6,0) arc (0:31:1.5);
        \draw[color=flatyellow, line width=0.8] (-0.4,-0.4) arc (180:110:1.1);

        \node at (0.4, 3.4) {$\left|0\right\rangle$};
        \node at (3.75,0) {$\left|1\right\rangle$};
        \node at (-0.9,-1.45) {$\left|2\right\rangle$};
        \node at (1.65,2.6) {$\left|\psi\right\rangle$};
        \node [color=flatred] at (0.25,1.15) {$\theta_0$};
        \node [color=flatblue] at (0.8,0.5) {$\theta_1$};
        \node [color=flatyellow] at (-0.4,0.45) {$\theta_2$};                    
    \end{tikzpicture}
    \caption{Schematic representation of a $3$-level qudit}
    \label{fig:threelevel}
\end{figure} 

We use an angle-based interpretation of qudit states to update them using rotation operations. Although the use of real amplitudes and the normalization condition reduce the degrees of freedom of a $q$-level qudit state to $q-1$, our representation uses $q$ angle values to simplify the formulation of rotation operators. A geometric interpretation of this representation for a $3$-level qudit is shown in Figure~\ref{fig:threelevel}. In this interpretation, the state vector lies in the non-negative section of a $(q-1)$-dimensional unit sphere embedded in a $q$-dimensional real amplitude space. The $i$-th angle $\theta_i$ is the angle between the state vector $\left| \psi \right\rangle$ and the basis vector $\left| i \right\rangle$, and is related to $c_i$ as follows: 

\begin{equation}
    c_i=\cos\theta_i, \quad \theta_i\in[0,\pi/2].
\end{equation}
Then, the normalization condition for a state becomes

\begin{equation}
    \sum_{i\in\boldsymbol{O}}\cos^2\theta_i = 1.
    \label{eq:multinormalization}
\end{equation}

As we show in Section~\ref{ssec:configuration}, the selected model structure determines the order in which qudits are measured. Regardless of the selected model, the measurement process in the MQEA variants is designed to produce only valid permutations. For this purpose, the outcomes obtained from previous measurements in the same solution are excluded from the following measurements. The probability of measuring outcome $i$ is expressed as

\begin{equation}
    p(i)=
    \begin{cases}
        \frac{\cos^2\theta_i}{\sum_{j\in\boldsymbol{A}}\cos^2\theta_j}, & \text{if}\;i\in\boldsymbol{A} \\
        0, & \text{otherwise},
    \end{cases}
    \label{eq:probability}
\end{equation}
where $\boldsymbol{A}\subseteq \boldsymbol{O}$ is the set of outcomes that have not been obtained from previous measurements. If the sum in the denominator is equal to zero, all $i\in\boldsymbol{A}$ are assigned equal selection probabilities. For a given qudit, excluding outcomes does not alter the relative probabilities of the allowed outcomes. In contrast, measurements in BQEA produces binary strings, so no outcome is excluded and $\boldsymbol{A}=\boldsymbol{O}=\{0,1\}$ for every measurement.

We use rotation operators to alter the probabilities of the outcomes for future iterations. The effect of the rotation operator $R(x,\Delta\theta)$, where $x$ is the target outcome and $\Delta\theta>0$ is the rotation magnitude, is shown by the equations below.

\begin{equation}
    \theta_{x}' = 
    \begin{cases}
        \theta_x-\Delta\theta, & \text{if}\;\theta_x-\Delta\theta > \theta_{\min}\\
        \theta_{\min}, & \text{otherwise}
    \end{cases}
    \label{eq:thetaxnew}
\end{equation}

\begin{equation}
    \theta_{i}' = \cos^{-1} \sqrt{ \cos^2\theta_i\frac{\sin^2\theta_x'}{\sin^2\theta_x}} ,\quad i\neq x
    \label{eq:thetainew}
\end{equation}

\begin{equation}
    \theta_i \gets \theta_{i}', \quad i\in\boldsymbol{O}
    \label{eq:thetaigets}
\end{equation}
Here, $\theta_{\min}$ is a configurable lower bound imposed during rotation operations to prevent complete convergence to a single outcome. As shown in Equation~\ref{eq:thetaxnew}, $R(x,\Delta\theta)$ aims to increase the probability of outcome $x$. It also adjusts the other angles according to  Equation~\ref{eq:thetainew}, so that the normalization condition and their relative measurement probabilities are preserved. If $\theta_x=\theta_{\min}$, no update is applied. Since the minimum angle is not enforced during the initialization step, it is possible that $\theta_x<\theta_{\min}$. In this case, the rotation operator sets $\theta_x\gets\theta_{\min}$ and assigns the other $\theta_i$ values equally:

\begin{equation}
    \theta_i\gets\cos^{-1}\frac{\sin\theta_{\min}}{\sqrt{q-1}},\quad i\neq x
\end{equation}

\subsection{Problem Encoding}

To encode a TSP instance with $n$ cities, we assign each city a unique index from the set $\{0,1,\ldots,n-1\}$. We fix city $0$ as the starting and ending location of each tour and represent the visiting order of the remaining cities using a permutation $(s_1,\ldots,s_{n-1})$, where $s_i\in\{1,\ldots,n-1\}$. Here, $s_i$ indicates that city $s_i$ is visited at the $(i+1)$-th position of the tour. In the MQEA variants, measurements produce valid permutations of $n-1$ integer values.

In our BQEA implementation, we use a qubit encoding based on assigning a unique integer index to each permutation, rather than encoding the city at each position of the tour independently. Under the TSP solution representation where city $0$ is fixed as the starting and ending point, our encoding requires $\lceil\log_2((n-1)!)\rceil$ qubits, compared with $(n-1)\lceil\log_2(n-1)\rceil$ qubits required by direct position-based encoding. However, there is still some redundancy as some binary strings may not represent valid permutations, since $2^{\lceil\log_2((n-1)!)\rceil}\geq(n-1)!$. In our implementation, invalid binary strings are assigned a large penalty cost, discouraging their selection as guiding solutions.

\subsection{Algorithm Structure}

\begin{figure*}[t]
    \centering
    \includegraphics[width=0.7\textwidth]{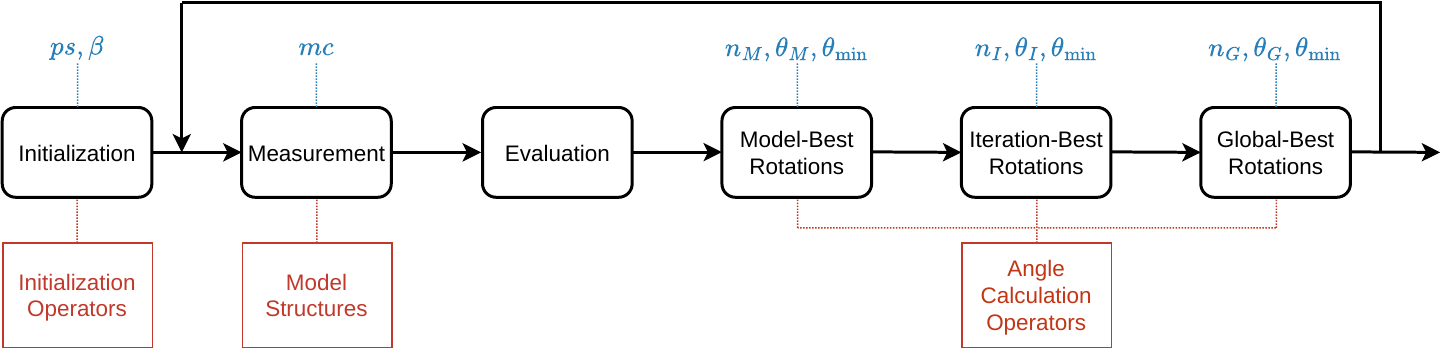}    
    \caption{The shared algorithm structure}
    \label{fig:qea}
\end{figure*}

Figure~\ref{fig:qea} shows the shared algorithm structure. The algorithm starts by initializing the probabilistic models in the population. It then executes multiple iterations, each starting with measurement operations to create candidate solutions. The selected model structure determines the order in which the qudits are measured and rotated. After measurement, the produced solutions are evaluated, and a fixed number of the best solutions for each model in the current iteration (model-best) is selected. Then, a fixed number of the best solutions produced across all models in the current iteration (iteration-best) are selected, and the list of the best solutions found so far (global-best) is updated. These lists are used to guide the model-best, iteration-best, and global-best rotations, respectively. Each rotation type is applied to every model a configurable number of times. The number of rotations for each type can also be set to zero, allowing a configuration to omit particular rotation types. When a guiding solution is used to update the model, the same rotation magnitude $\Delta\theta$ is used in the rotation operators for all modified qudits. The $\Delta\theta$ value for a rotation is calculated by the selected angle calculation operator according to the rotation type and the associated base rotation angle. Iterations continue until a termination condition is satisfied. Since we consider time-limited optimization, the algorithm terminates when the estimated duration of the next iteration would cause the total execution time to exceed the specified time-limit. This estimate is based on the average duration of completed iterations. After termination, the best solution found so far is returned.

\subsection{Configuration Space}
\label{ssec:configuration}

Figure~\ref{fig:qea} shows the hyperparameters (in blue) and the set of configuration components (in red) of the configuration space, which we describe below.

\subsubsection{Hyperparameters}

The population size $ps$ determines the number of probabilistic models, and the measurement count $mc$ determines the number of candidate solutions produced from each model in each iteration. Six hyperparameters are used to control the influence of the three rotation types. The parameters $n_M$, $n_I$, and $n_G$ indicate the numbers of model-best, iteration-best, and global-best solutions used to guide the rotations. Since one rotation is applied for each guiding solution, these parameters also determine the number of rotations of each type applied to each model in each iteration. Similarly, $\theta_M$, $\theta_I$, and $\theta_G$ indicate the base rotation angles used in model-best, iteration-best, and global-best rotations. These values serve as parameters for the selected angle calculation operator to determine the actual rotation angles. As we explain in Section~\ref{ssec:qudit}, $\theta_{\min}$ defines the lower bound imposed during rotation operations. Finally, $\beta$ determines the influence of problem-specific heuristic information during heuristic initialization.

\subsubsection{Model Structures}

We implement two model structures that determine the order in which qudits are accessed during measurement and rotation operations. The model structures are explained below.

\begin{itemize}[itemsep=1mm, leftmargin=0mm, label={}]

    \item (1) \textbf{Sequential Model:} In this model, each qudit represents a fixed position in the solution representation. During measurement and rotation operations, qudits are accessed in a fixed order. BQEA and baseline MQEA use the sequential model. For an $n$-city TSP, the BQEA model consists of $\lceil \log_2((n-1)!)\rceil$ qubits, each representing a binary value in the permutation index encoding. For baseline MQEA the sequential model consists of $n-1$ qudits, each providing an integer in $\{1,\ldots,n-1\}$. During model update, $R(x_i,\Delta\theta)$ is applied to the $i$-th qudit, if the $i$-th value of the guiding solution is $x_i$.

    \item (2) \textbf{Graph-Based Model:} In this model, each qudit represents a city rather than a fixed position in a permutation. MQEA-G and MQEA-GH use this model. The graph-based model for an $n$-city problem consists of $n$ qudits, each of which is assigned a city index in $\{0,\ldots,n-1\}$. At the beginning of the measurement process, qudit $0$ is measured, providing the first integer. Each measurement produces an integer in $\{1,\ldots,n-1\}$, which indicates the index of the next measured qudit and provides an integer for the solution permutation. The same ordering procedure is used for rotation operations applied to a model. First, $R(s_1,\Delta\theta)$ is applied to qudit $0$. Then, for each $i\in\{2,\ldots,n-1\}$, $R(s_i,\Delta\theta)$ is applied to qudit $s_{i-1}$. 

    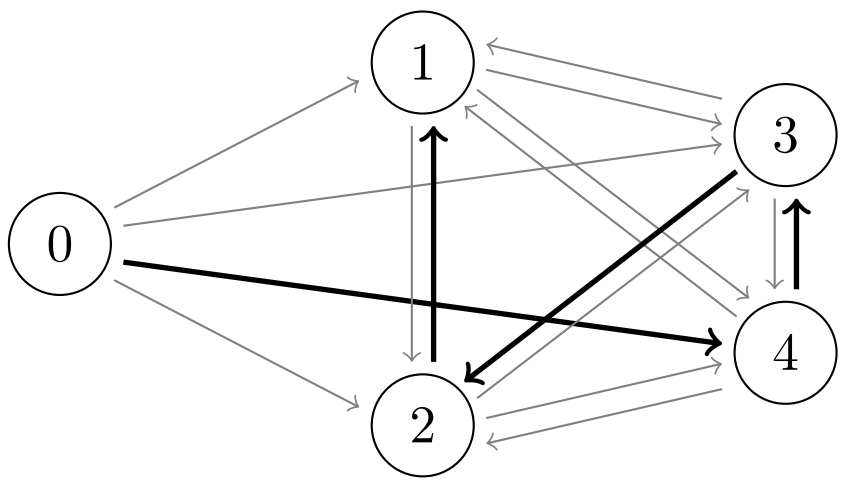
\begin{figure}
        \centering
        \begin{tikzpicture}[scale=1.2]
        \draw (0,0) circle (8pt);
        \draw (2,1) circle (8pt);
        \draw (2,-1) circle (8pt);
        \draw (4,-0.6) circle (8pt);
        \draw (4,0.6) circle (8pt);

        \node at (0,0) {0};
        \node at (2, 1) {1};
        \node at (2, -1) {2};
        \node at (4, 0.6) {3};
        \node at (4, -0.6) {4};

        \draw[->, color=gray] (0.3,0.2) -- (1.65,0.9);
        \draw[->, color=gray] (0.3,-0.2) -- (1.65,-0.9);
        \draw[->, color=gray] (0.35,0.1) -- (3.65,0.55);
        \draw[->, line width=1] (0.35,-0.1) -- (3.65,-0.55);

        \draw[->, color=gray] (1.94,0.65) -- (1.94,-0.65);
        \draw[->, line width=1] (2.06,-0.65) -- (2.06,0.65);

        \draw[->, color=gray] (3.94,0.25) -- (3.94,-0.25);
        \draw[->, line width=1] (4.06,-0.25) -- (4.06,0.25);

        \draw[->, color=gray] (2.35,0.96) -- (3.65,0.66);
        \draw[->, color=gray] (3.65,0.8) -- (2.35,1.1);

        \draw[->, color=gray] (2.35,-0.96) -- (3.65,-0.66);
        \draw[->, color=gray] (3.65,-0.8) -- (2.35,-1.1);

        \draw[->, color=gray] (2.3,0.85) -- (3.8,-0.3);
        \draw[->, color=gray] (3.73,-0.4) -- (2.23,0.76);

        \draw[->, color=gray] (2.3,-0.85) -- (3.8,0.3);
        \draw[->, line width=1] (3.73,0.4) -- (2.23,-0.76);
                
    \end{tikzpicture}
    \caption{An example measurement on the graph-based model}
    \label{fig:graphbased}
    \end{figure}

    Figure~\ref{fig:graphbased} shows an example tour on a graph-based model for a TSP with $5$ cities. Each dark arrow shows the outcome measured from a qudit. The permutation obtained from this measurement is $4\;3\;2\;1$, which represents the tour $0\rightarrow4\rightarrow3\rightarrow2\rightarrow1\rightarrow0$.

\end{itemize}

\subsubsection{Initialization Operators}

The two initialization operators implemented in the configuration space are explained below.

\begin{itemize}[itemsep=1mm, leftmargin=0mm, label={}]

    \item (1) \textbf{Standard Initialization:} This operator initializes all qudits in all probabilistic models to a uniform superposition. For a problem with $n$ cities, the state of each qudit is initialized to

    \begin{equation}
        \left|\psi\right\rangle=\sum_{i\in\boldsymbol{O}} \frac{1}{\sqrt{|\boldsymbol{O}|}}\left|i\right\rangle,
    \end{equation}
    where $\boldsymbol{O}=\{0,1\}$ for BQEA and $\boldsymbol{O}=\{1,\ldots,n-1\}$ for the MQEA variants. In this state, all outcomes have equal measurement probabilities. Standard initialization is used in all configurations of BQEA, baseline MQEA, and MQEA-G.

    \item (2) \textbf{Heuristic Initialization:} The heuristic initialization method is designed to be used together with the graph-based model, and it uses the inverse-distance heuristic commonly used in ACO. Although heuristic initialization could also be implemented for a sequential model using a suitable heuristic, we do not consider this option in this study.
    
    In heuristic initialization, the weight of outcome $i$ from qudit $j$ is calculated as

    \begin{equation}
        w_{ji}=
        \begin{cases}
            \left(\frac{1}{d_{ji}}\right)^\beta, & \text{if } i\neq j \\
            0, & \text{if } i=j
        \end{cases}        
    \end{equation}
    where $d_{ji}$ is the distance between cities $i$ and $j$, and $\beta$ is the heuristic influence. Due to the normalization condition, the initial state becomes

    \begin{equation}
        \left|\psi_j\right\rangle=\sum_{i\in\boldsymbol{O}}\sqrt{\frac{w_{ji}}{\sum_{k\in\boldsymbol{O}}w_{jk}}}\left|i\right\rangle.
    \end{equation}
    In the angle-based representation of this state, the angle associated with the $i$-th outcome of the $j$-th qudit is

    \begin{equation}
        \theta_{ji}=\cos^{-1}\sqrt{\frac{w_{ji}}{\sum_{k\in\boldsymbol{O}}w_{jk}}}
    \end{equation}

    Heuristic initialization is used for MQEA-GH configurations. Although the heuristic information used is the same as that commonly used in ACO, its integration differs. In ACO, heuristic information influences the construction of each solution, whereas MQEA-GH utilizes heuristic information only during initialization.
    
\end{itemize}

\subsubsection{Angle-Calculation Operators}

Our implementation provides three angle-calculation operators, one of which is selected for each BQEA, baseline MQEA, MQEA-G, and MQEA-GH configuration. The angle-calculation operators are described below.

\begin{itemize}[itemsep=1mm, leftmargin=0mm, label={}]

    \item (1) \textbf{Constant Angle:} No additional calculation is applied, and $\theta_M$, $\theta_I$, and $\theta_G$ are used for every model-best, iteration-best, and global-best rotation, respectively.

    \item (2) \textbf{Cost-Based Calculation:} The cost of the guiding solution is used in the calculation of the rotation angle. Rotation angles are calculated as
    
    \begin{equation}
        \begin{split}
            \Delta\theta=\theta_M\frac{c_b}{c}, & \text{ for a model-best rotation}, \\
            \Delta\theta=\theta_I\frac{c_b}{c}, & \text{ for an iteration-best rotation}, \\
            \Delta\theta=\theta_G\frac{c_b}{c}, & \text{ for a global-best rotation}, \\
        \end{split}         
    \end{equation}
    where $c$ is the cost of the guiding solution, and $c_b$ is the baseline cost obtained by evaluating a random solution at the beginning of the algorithm.

    \item (3) \textbf{Rank-Based Calculation:} For each rotation type, guiding solutions are ranked according to their costs, with the best solution receiving rank $0$. The rotation angles are calculated as

    \begin{equation}
        \begin{split}
            \Delta\theta=\theta_M\frac{n_M-r}{n_M}, &\text{ for a model-best rotation}, \\
            \Delta\theta=\theta_I\frac{n_I-r}{n_I}, & \text{ for an iteration-best rotation}, \\
            \Delta\theta=\theta_G\frac{n_G-r}{n_G}, & \text{ for a global-best rotation}, \\
        \end{split}         
    \end{equation}
    where $r$ is the rank of the solution.    
\end{itemize}

\section{Experimental Evaluation}
\label{sec:experimental}

We performed an experimental evaluation using a large number of configurations of BQEA, MQEA, MQEA-G, MQEA-GH, ACO, GA, and SA. The evaluation has three main objectives. The first is to investigate the impact of the qudit representation, graph-based model, and heuristic initialization under strict computational budgets by comparing the performance of BQEA, MQEA, MQEA-G, and MQEA-GH. The second is to assess the performance of MQEA-GH relative to the established metaheuristics ACO, GA, and SA under various computational budgets. The third is to show how the preferred operators and hyperparameter values of MQEA-GH vary across time limits and instance characteristics. In the following, we first introduce the experimental setup and then present the analysis results.

\subsection{Experimental Setup}

We considered 191 time limits from 50 to 1000 ms in increments of 5 ms. To evaluate configurations under various time limits, we used the analysis tool provided by the STILO framework. Rather than requiring a separate execution for each time limit, the tool records the best objective value obtained before each specified time limit during a single execution. This also allows us to evaluate configurations under strict time limits without relying on estimates of iteration duration. We conducted the experiments on a virtual machine running Ubuntu 24.04 LTS and hosted using Proxmox VE 9.2 on a Supermicro server equipped with an AMD EPYC 9354 processor and 192 GB of RAM. The virtual machine was allocated 62 vCPUs and 64 GB of RAM. We evaluated 60 configurations concurrently on separate threads to reduce the total duration of the experiments. As a more consistent measure in a parallel computing environment, we measured per-thread CPU time rather than wall-clock time.

We generated the configurations of each algorithm family using different combinations of parameter values and operators. For GA and SA, we used the configurations evaluated in \cite{ccalikyilmaz2026metaheuristic}, and do not repeat their parameter values and components here. The parameter values used to generate the configurations of the other algorithms are given below. All valid and non-redundant combinations of these values were used in the experiments. Invalid ones, such as those with all rotation counts equal to zero, and redundant ones, such as those with differing base angle values for a certain rotation type whose rotation count is zero, are discarded.

\begin{itemize}[itemsep=1mm, leftmargin=0mm, label={}]
    \item \textbf{BQEA/MQEA/MQEA-G Configurations:} For all three algorithms, two population sizes ($ps\in\{4,8\}$), two measurement counts ($mc\in\{10,20\}$), and three limit angles ($\theta_{\min}\in\{0,0.05,0.1\}$ radians) were used. For each rotation type, four rotation counts and five base rotation angles ($n_M,n_I,n_G\in\{0,1,2,3\}$ and $\theta_M,\theta_I,\theta_G\in\{0.001,0.0025,0.005,0.01,0.02\}$ radians) were considered. In addition, three different angle-calculation operators were used. The configurations of these three algorithms differ, as BQEA uses a sequential qubit model, MQEA uses a sequential qudit model, and MQEA-G uses a graph-based qudit model. All these algorithms use standard initialization.
    
    \item \textbf{MQEA-GH Configurations:} MQEA-GH configurations use a graph-based qudit model and heuristic initialization. The parameter values used for MQEA-GH configurations are the same as those used for BQEA, MQEA, and MQEA-G, with the addition of seven heuristic influence values ($\beta\in\{1,2,3,5,10,15,20\}$).

    \item \textbf{ACO Configurations:} Similar to GA and SA, we based the ACO configurations on those given in \cite{ccalikyilmaz2026metaheuristic}, with two modifications. First, we considered only the single-stage graph structure, rather than the two structures evaluated in the referenced study. Second, we extended the set of heuristic influence values to $\beta\in\{1,2,3,5,10,15,20\}$. The purpose of these changes is to improve the comparability of ACO and MQEA-GH configurations by using similar graph-based representations and the same heuristic influence values.
\end{itemize}

To investigate how algorithm performance depends on instance characteristics, we generated synthetic instances with randomly sampled distance values. For this purpose, we defined four instance classes by combining two city counts ($\{50,100\}$) and two normal distributions with a fixed mean ($\mu=100$) and different standard deviations ($\sigma\in\{10,25\}$). For each instance class, we generated 10 random instances by independently sampling the distance for each undirected edge, resampling any negative values. The resulting instances are symmetric but not necessarily metric. We executed each configuration once on every instance and used the average objective value across the 10 instances as the measure of solution quality. For each instance class and time limit, we identified the configuration achieving the lowest average objective value within each algorithm family. We used these configurations to compare the best observed performance of the algorithm families.

We complemented the synthetic experiments with evaluations on four established benchmark instances from TSPLIB~\cite{reinelt1991tsplib}: berlin52, eil76, rd100, and tsp225. For each benchmark instance, we performed 10 independent executions of every configuration and recorded the best objective value achieved before each of the 191 time limits. We used the average objective value across these executions as the measure of solution quality. Similar to the synthetic analysis, we identified the configuration within each algorithm family achieving the lowest average objective value for each benchmark instance and time limit, and compared the best observed performance of the algorithm families.

\subsection{Comparison of the Algorithm Families}

\begin{figure*}[!t]
    \centering
    \begin{subfigure}{0.32\textwidth}
        \includegraphics[width=\textwidth]{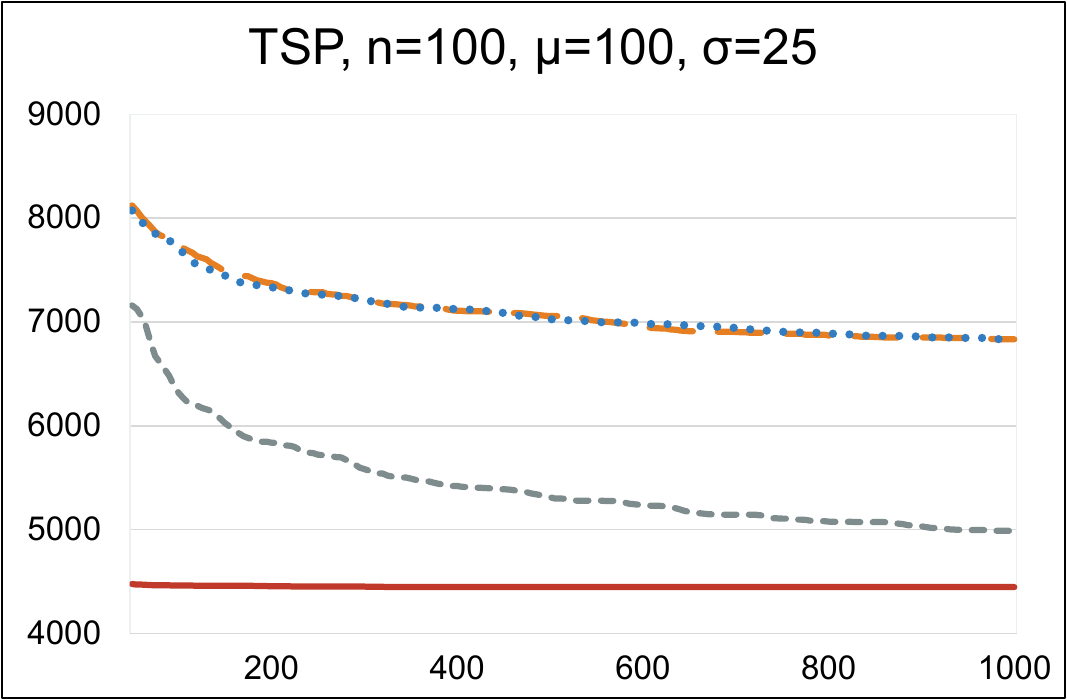}
    \end{subfigure}
    \hfill
    \begin{subfigure}{0.32\textwidth}
        \includegraphics[width=\textwidth]{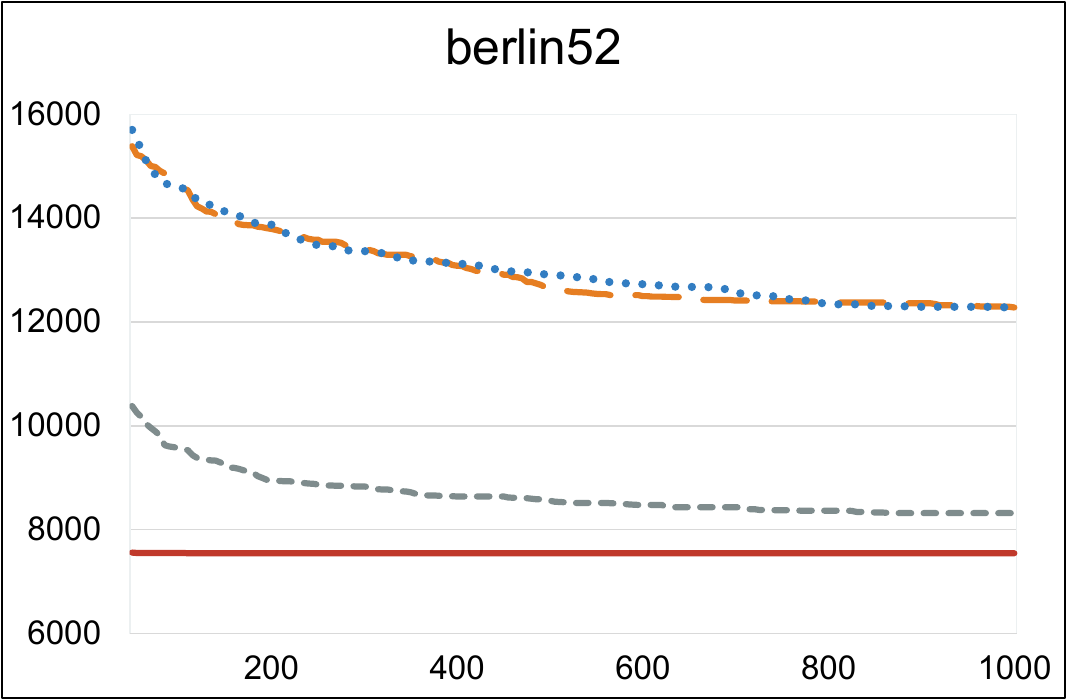}
    \end{subfigure}
    \hfill
    \begin{subfigure}{0.32\textwidth}
        \includegraphics[width=\textwidth]{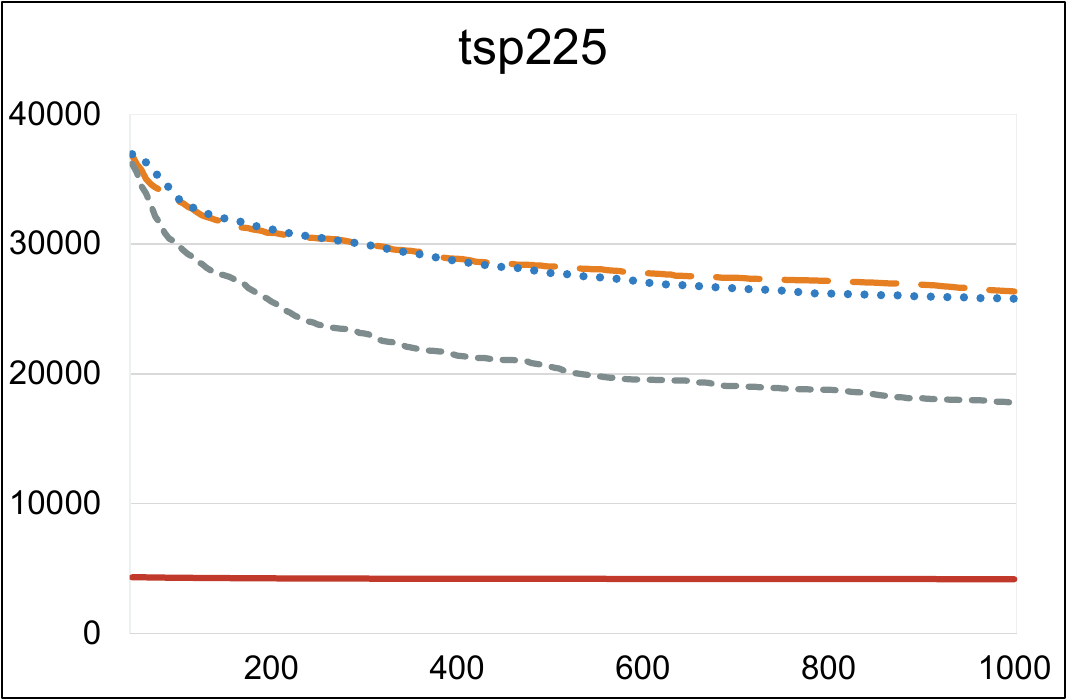}
    \end{subfigure}
    
    \caption{Comparison of the QEA variants across time limits (ms). The lines show the average objective values obtained by the best-performing configurations of each algorithm family at each time limit: {\color{flatorange} BQEA (orange long-dashed)}, {\color{flatblue} MQEA (blue dotted)}, {\color{flatgray} MQEA-G (gray short-dashed)}, and {\color{flatred} MQEA-GH (red solid)}.}
    \label{fig:qeacomp}
\end{figure*}

As the first step of the experimental analysis, we compared the performance of the best-performing configurations of BQEA, MQEA, MQEA-G, and MQEA-GH. For all tested instance classes and benchmark instances, the best-performing configurations of MQEA-G outperformed those of BQEA and MQEA under all time limits, with substantial differences for most cases. Similarly, best-performing MQEA-GH configurations achieved considerably lower objective values than MQEA-G configurations. These results demonstrate the effectiveness of the two proposed mechanisms. The graph-based qudit model substantially improves the solution quality compared with the sequential qudit model, and heuristic initialization improves it further. Another noteworthy finding is that MQEA did not consistently outperform BQEA over the entire time-limit range for any tested setting. Both algorithms performed considerably worse than the other two variants, while their relative performance depended on the instance and time limit. This result shows that, for the tested instances, a qudit representation without a graph-based model and heuristic information did not consistently outperform a qubit representation with a permutation-index encoding. Figure~\ref{fig:qeacomp} presents representative comparisons of the QEA variants.

\begin{figure*}[!t]
    \centering
    \begin{subfigure}{0.32\textwidth}
        \includegraphics[width=\textwidth]{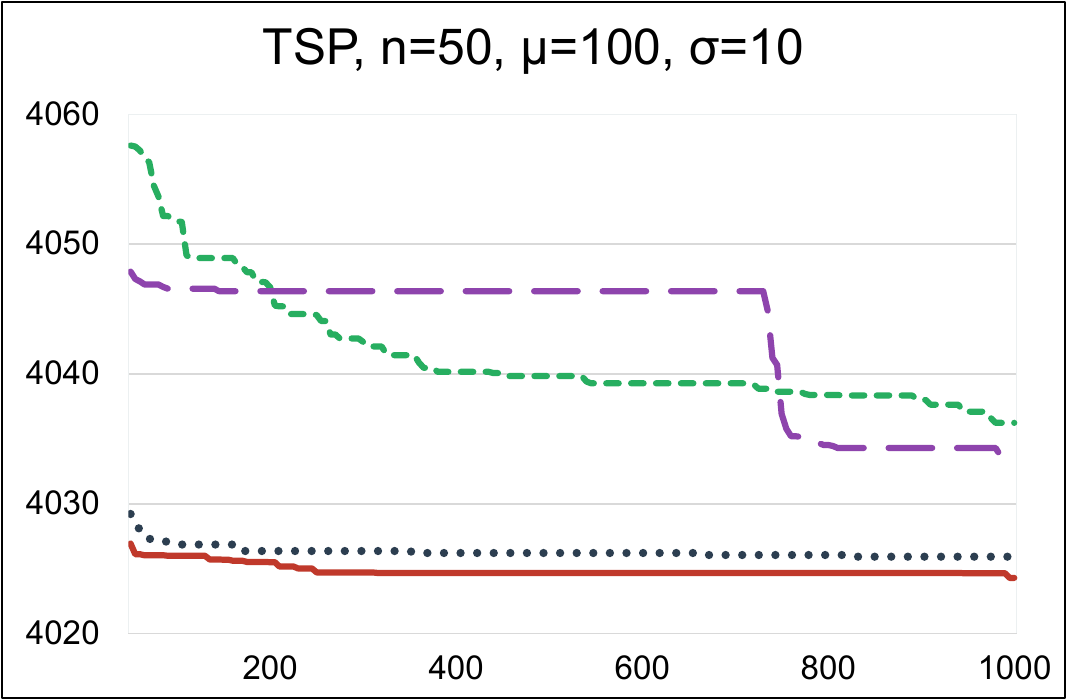}
    \end{subfigure}
    \hfill
    \begin{subfigure}{0.32\textwidth}
        \includegraphics[width=\textwidth]{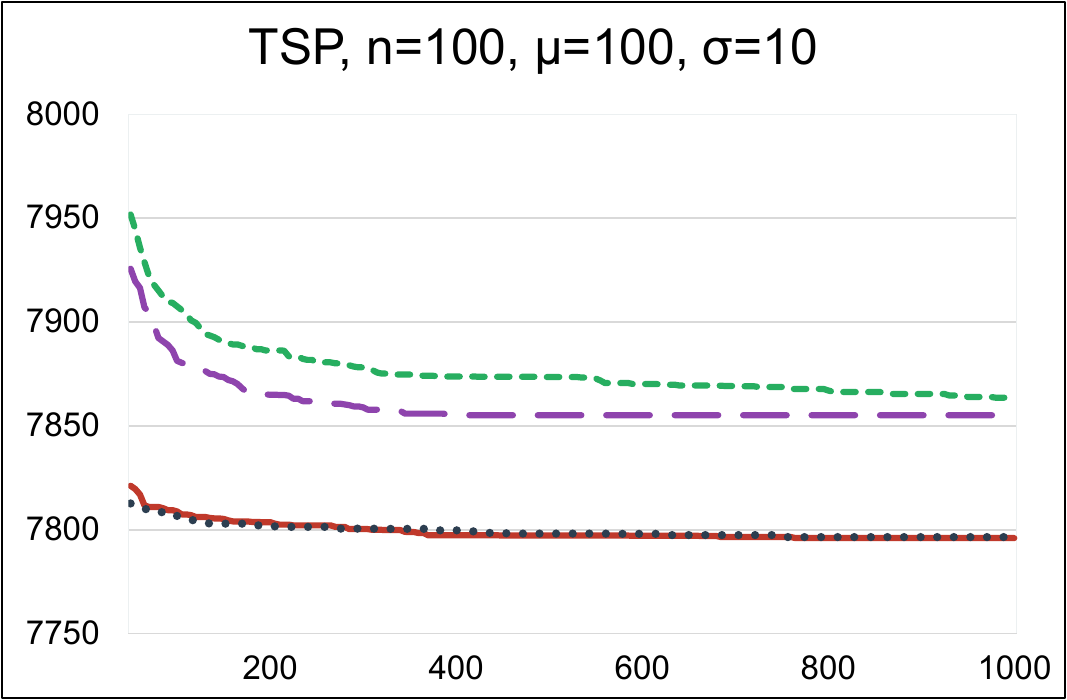}
    \end{subfigure}
    \hfill
    \begin{subfigure}{0.32\textwidth}
        \includegraphics[width=\textwidth]{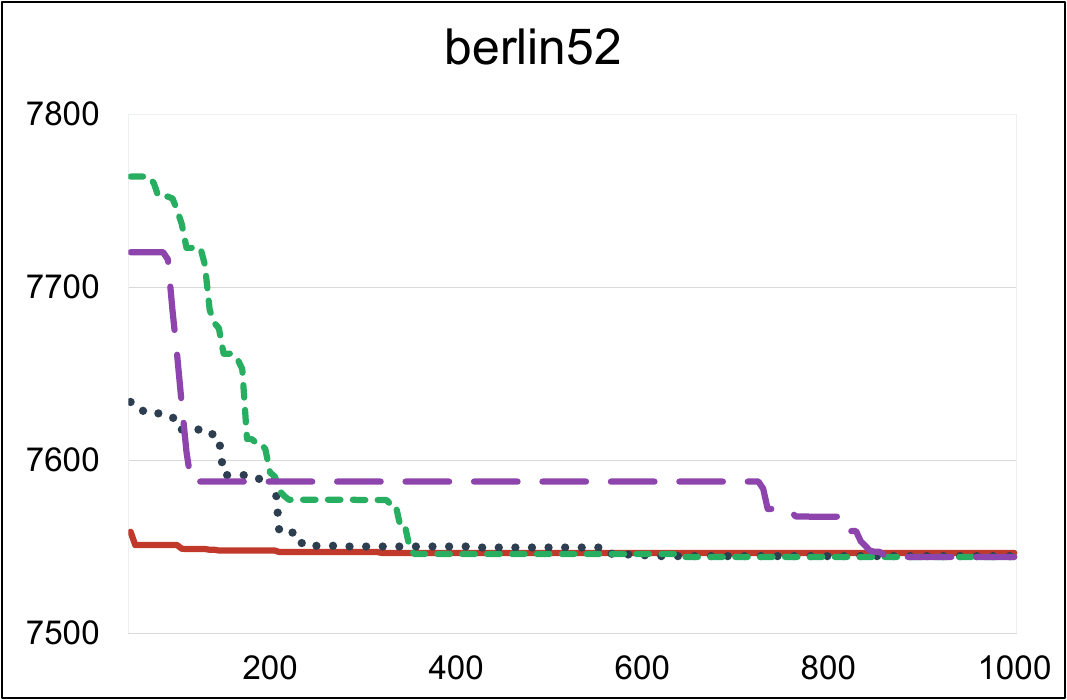}
    \end{subfigure}
    \begin{subfigure}{0.32\textwidth}
        \includegraphics[width=\textwidth]{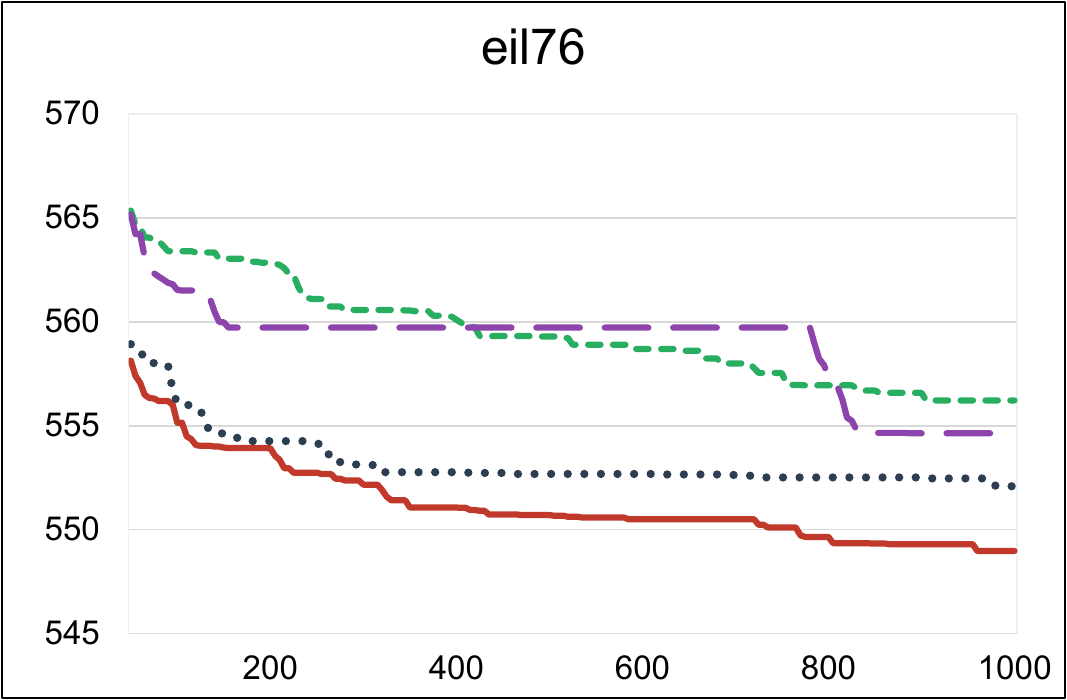}
    \end{subfigure}
    \hfill
    \begin{subfigure}{0.32\textwidth}
        \includegraphics[width=\textwidth]{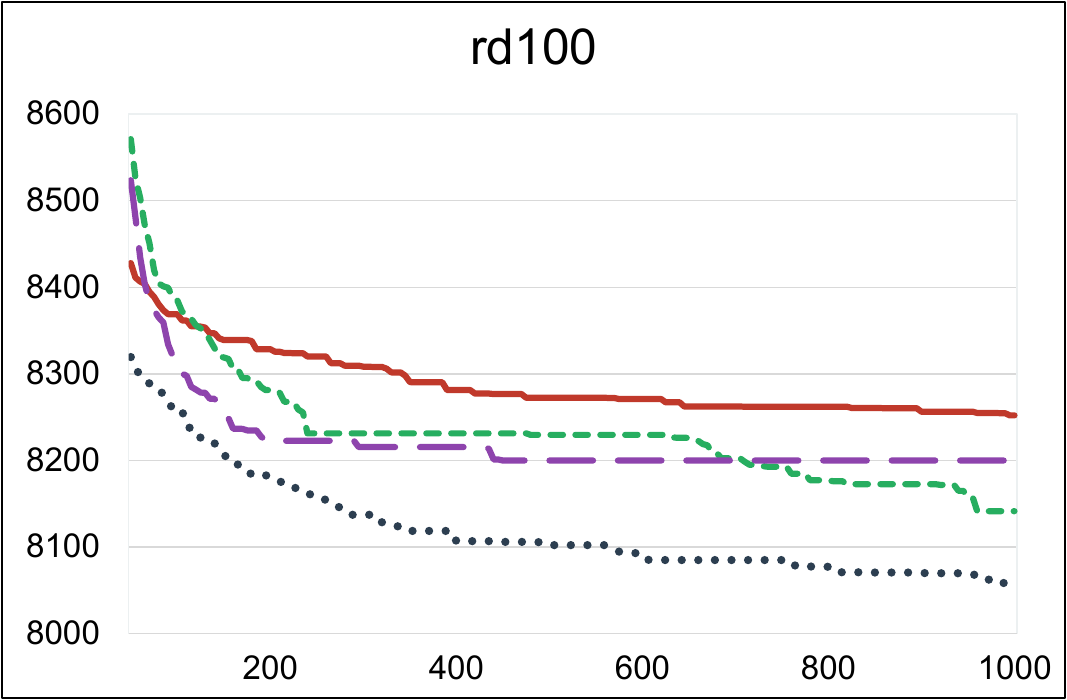}
    \end{subfigure}
    \hfill
    \begin{subfigure}{0.32\textwidth}
        \includegraphics[width=\textwidth]{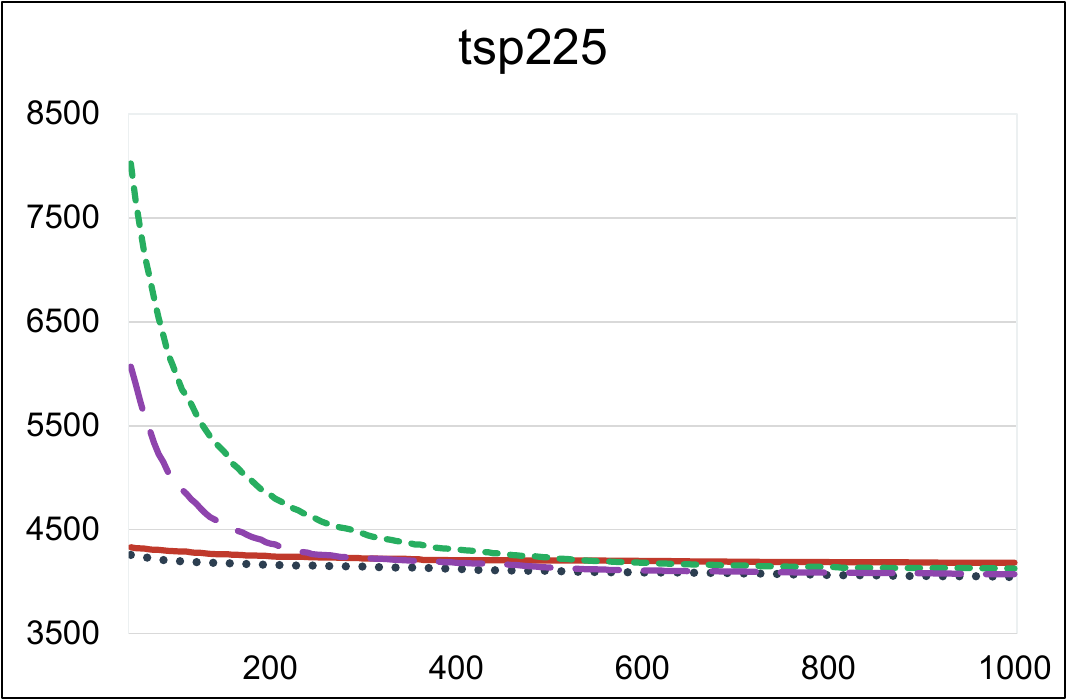}
    \end{subfigure}
    
    \caption{Comparison of MQEA-GH with established metaheuristics across time limits (ms). The lines show the average objective values obtained by the best-performing configurations of each algorithm family at each time limit: {\color{flatlaci} ACO (dark-blue dotted)}, {\color{flatgreen} GA (green short-dashed)}, {\color{flatpurple} SA (purple long-dashed)}, and {\color{flatred} MQEA-GH (red solid)}.}
    \label{fig:metacomp}
\end{figure*}

Given its substantial lead in the achieved objective values among the QEA variants, we only compared MQEA-GH with established metaheuristics. The results show that the best-performing MQEA-GH configurations were generally competitive with ACO, GA, and SA for tested instances. For all the synthetic instance classes, MQEA-GH surpassed GA and SA under all tested time limits. It also achieved equal or lower objective values than ACO for the classes with $n=50,\;\sigma=10$ and $n=100,\;\sigma=25$ under all time limits. For the other two instance classes, ACO and MQEA-GH showed very similar results, with their relative performance alternating across time limits. MQEA-GH was not as dominant for the benchmark instances, except for eil76, for which the best-performing MQEA-GH configurations achieved the best solution quality throughout the evaluated time-limit range. MQEA-GH also showed the best performance for berlin52 under short time limits, though the other algorithm families caught up with it under longer limits. It also achieved the second-lowest average objective values for the short time limits of tsp225, but GA and SA configurations were able to match its performance and even slightly surpass it under larger computational budgets. MQEA-GH was the worst-performing algorithm for rd100, as it obtained the largest objective values for most of the time-limit range.

Overall, MQEA-GH was among the two best-performing algorithm families for all synthetic instance classes, while its performance on benchmark instances was more dependent on instance characteristics and computational budget. Across most tested instances, MQEA-GH was particularly effective under small computational budgets, obtaining high-quality solutions within very short time limits. However, the objective values found by the best MQEA-GH configurations showed relatively limited improvement as the computational budget increased for some instances. These observations indicate that MQEA-GH is especially useful for optimization under strict time limits, although its relative performance may decline for some instances when larger computational budgets are available. Figure~\ref{fig:metacomp} shows representative results comparing MQEA-GH configurations with those of established metaheuristics.

\subsection{Analysis of the Best-Performing MQEA-GH Configurations}

\begin{figure*}[!t]
    \centering
    \begin{subfigure}{0.32\textwidth}
        \includegraphics[width=\textwidth]{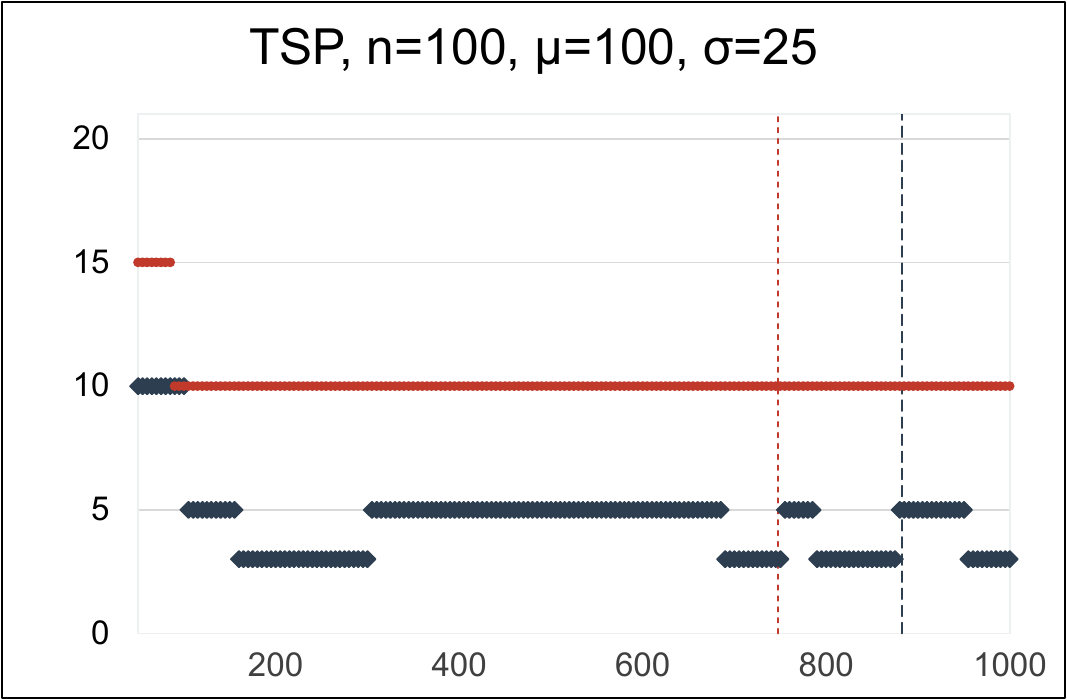}
    \end{subfigure}
    \hfill
    \begin{subfigure}{0.32\textwidth}
        \includegraphics[width=\textwidth]{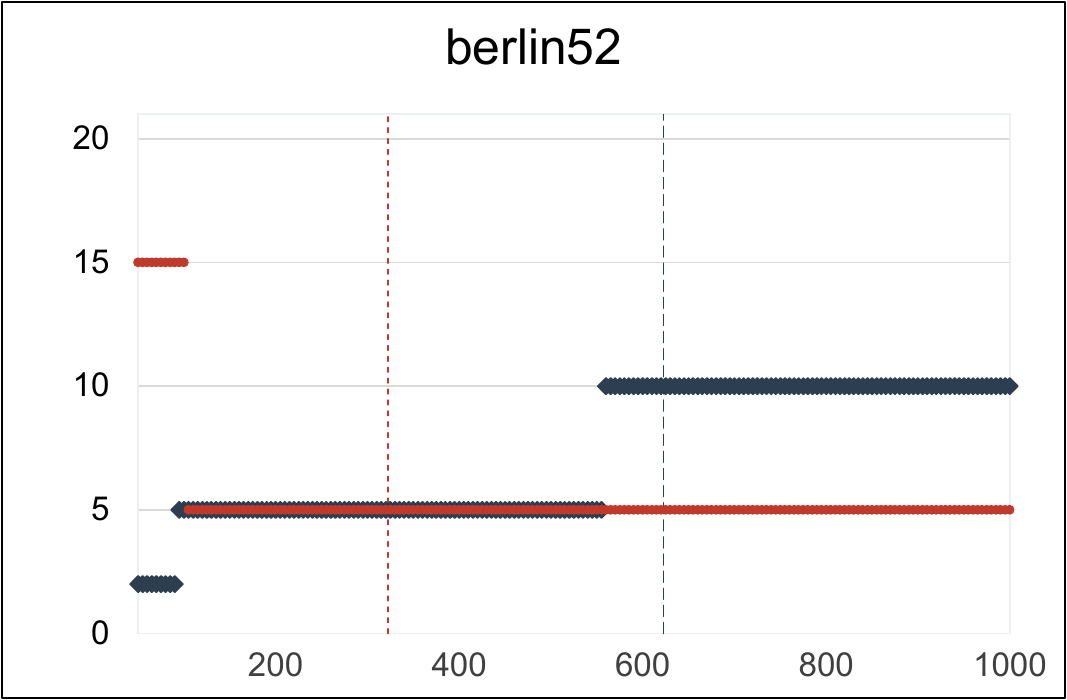}
    \end{subfigure}
    \hfill
    \begin{subfigure}{0.32\textwidth}
        \includegraphics[width=\textwidth]{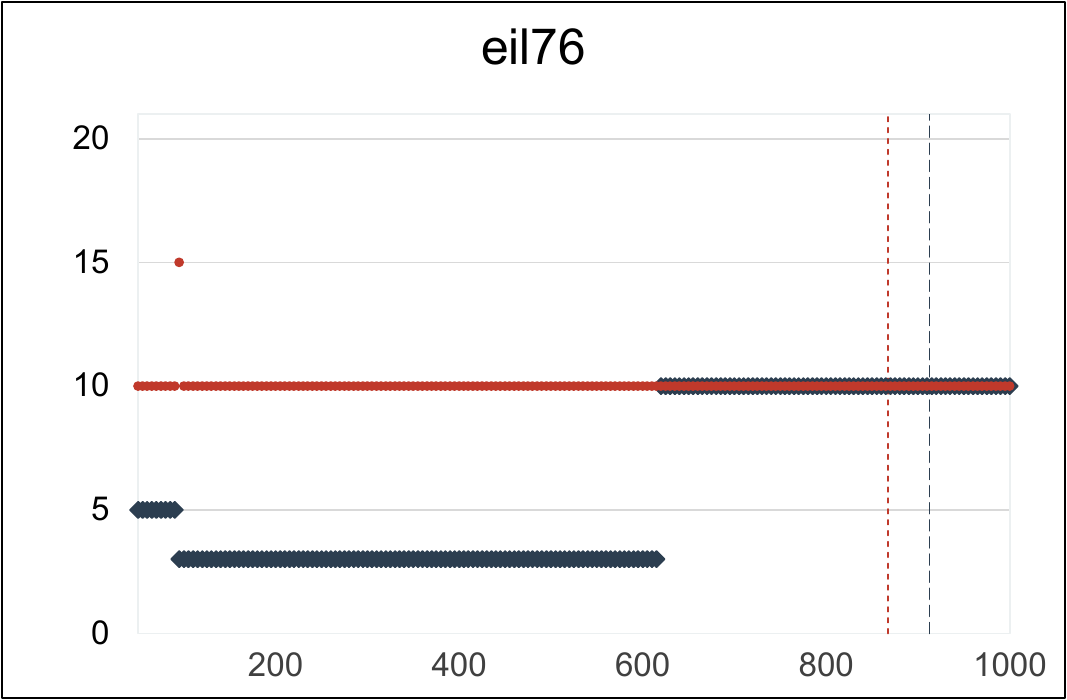}
    \end{subfigure}
    
    \caption{Heuristic influence values selected by the best-performing {\color{flatlaci} ACO (dark-blue diamonds)} and {\color{flatred} MQEA-GH (red disks)} configurations at each time limit (ms). The {\color{flatlaci} dark-blue long-dashed} and {\color{flatred} red short-dashed} vertical lines show the stabilization time limits of ACO and MQEA-GH configurations, respectively.}
    \label{fig:heuristic}
\end{figure*}

\begin{figure*}[!t]
    \centering
    \begin{subfigure}{0.32\textwidth}
        \includegraphics[width=\textwidth]{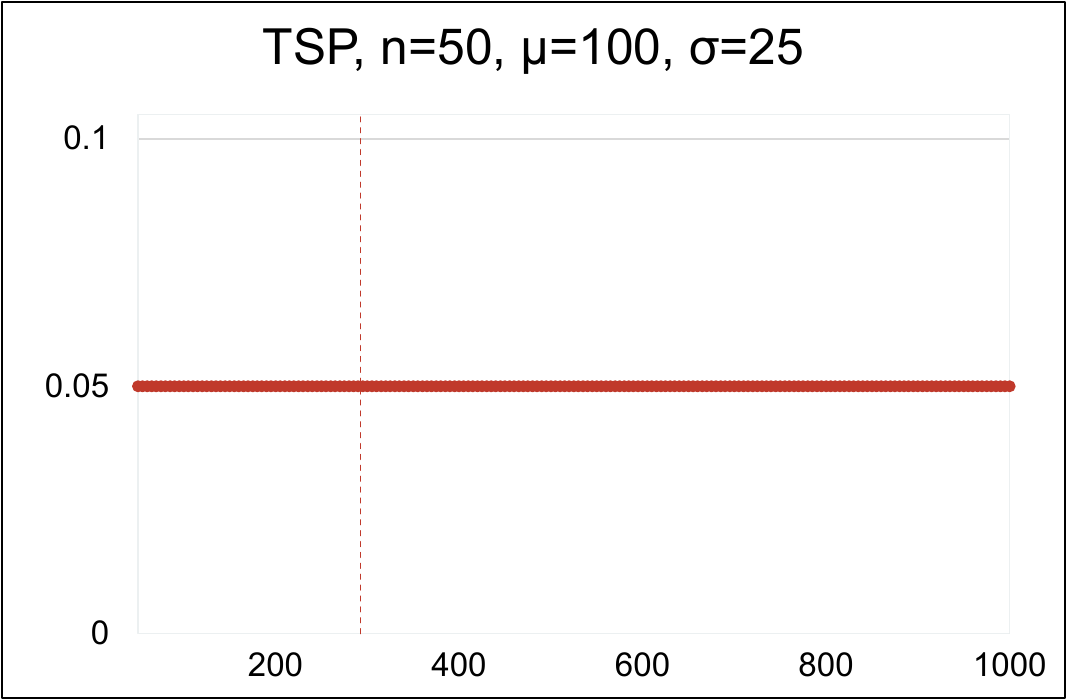}
    \end{subfigure}
    \hfill
    \begin{subfigure}{0.32\textwidth}
        \includegraphics[width=\textwidth]{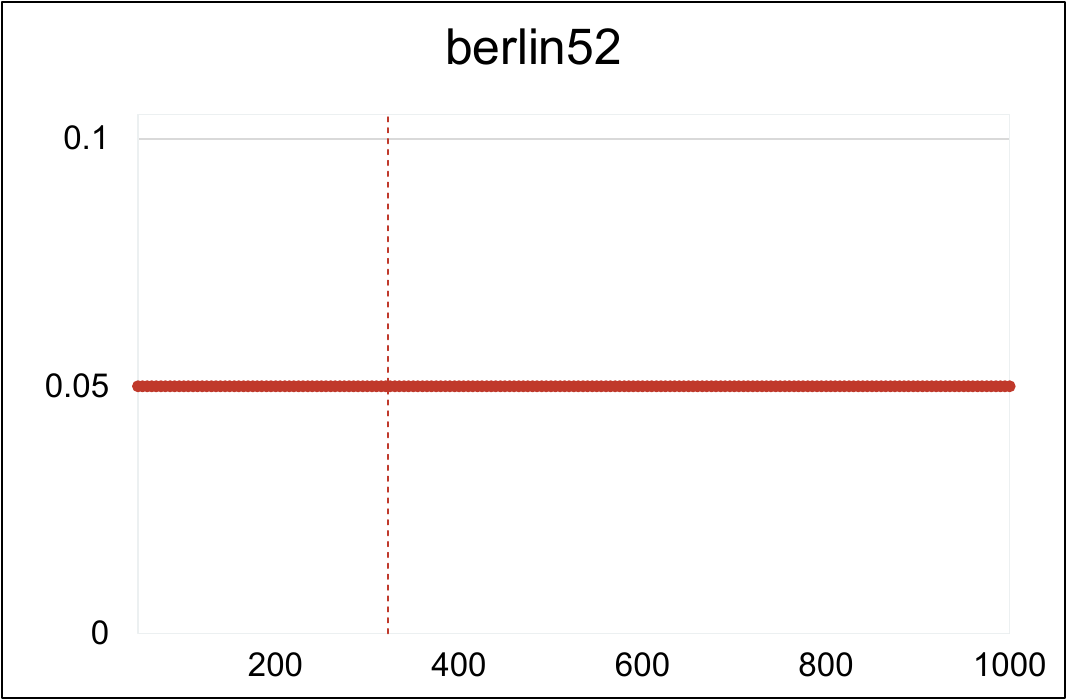}
    \end{subfigure}
    \hfill
    \begin{subfigure}{0.32\textwidth}
        \includegraphics[width=\textwidth]{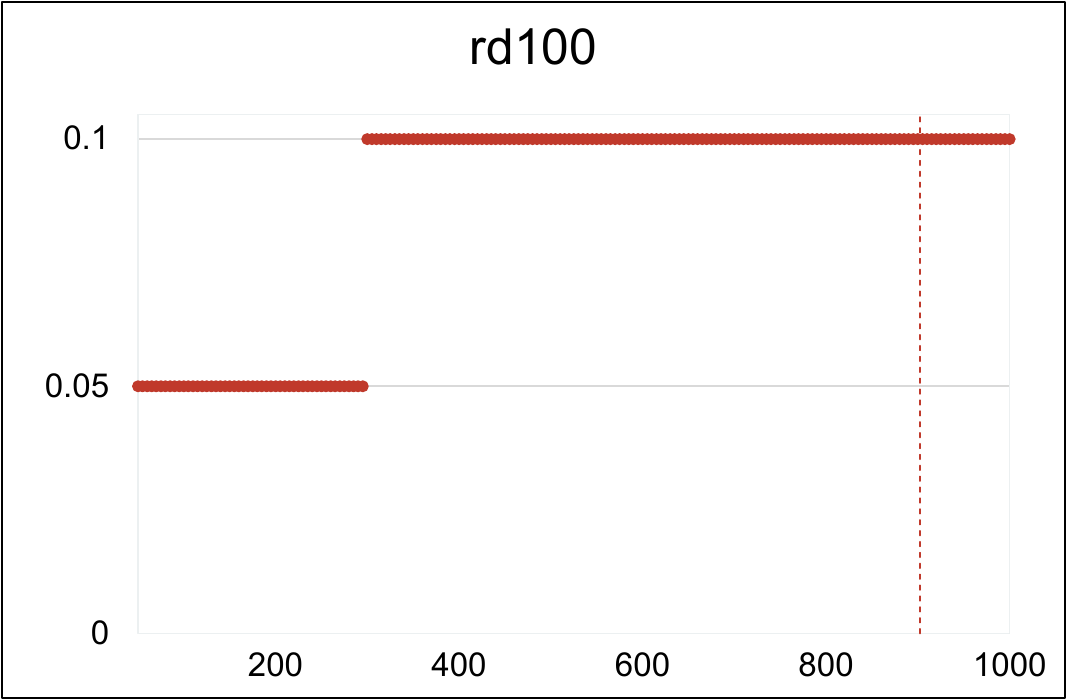}
    \end{subfigure}
    
    \caption{Limit angle values selected by the best-performing MQEA-GH configurations at each time limit (ms). Vertical lines indicate the stabilization time limits.}
    \label{fig:limit}
\end{figure*}

In this subsection, we analyze the hyperparameter values and operators selected by the best-performing MQEA-GH configurations across tested instance classes, benchmark instances, and time limits. We define the stabilization time limit as the largest time limit at which the best observed value improves. Since multiple configurations are more likely to achieve similar objective values after stabilization, the portions before stabilization time limits are given greater emphasis. Due to space considerations, we omit the graphs for some settings and include only representative cases for each analyzed configuration component.

Heuristic influence values selected by the best-performing ACO and MQEA-GH configurations show that generally higher numerical values are preferred for MQEA-GH. This is potentially caused by the different mechanisms used by the algorithms to incorporate heuristic information. Since MQEA-GH uses heuristic information only during initialization, higher heuristic influence values may be less likely to cause premature convergence. The main exceptions are certain portions of the time-limit ranges for the instance class with $n=50,\;\sigma=10$ and berlin52, where higher values were preferred for ACO. Another important observation is that the heuristic influence selection for MQEA-GH is relatively stable across time limits. Generally, a single value is preferred for each instance class or benchmark instance, with only minor variations across time limits. This suggests that the preferred heuristic influence value for MQEA-GH mostly depends on instance characteristics rather than the available computational budget. Figure~\ref{fig:heuristic} illustrates these results.

For all tested instances and time limits, the best-performing configurations preferred positive limit angles, showing a persistent preference for constraining qudit amplitudes. For synthetic instance classes, a limit angle of $0.05$ radians was preferred for almost all time limits. Benchmark instances exhibited more varied preferences, as both values were selected for different portions of the time-limit range. The only exception was berlin52, for which a limit angle of $0.05$ radians was consistently selected. Figure~\ref{fig:limit} presents these results.

\begin{figure*}[!t]
    \centering
    \begin{subfigure}{0.32\textwidth}
        \includegraphics[width=\textwidth]{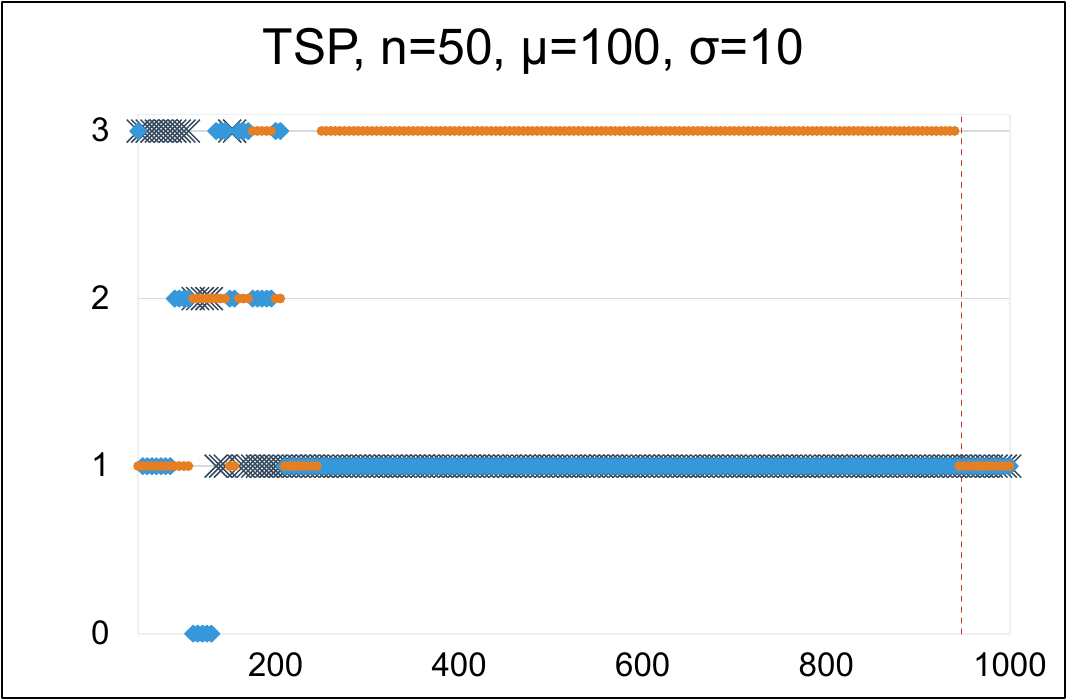}
    \end{subfigure}
    \hfill
    \begin{subfigure}{0.32\textwidth}
        \includegraphics[width=\textwidth]{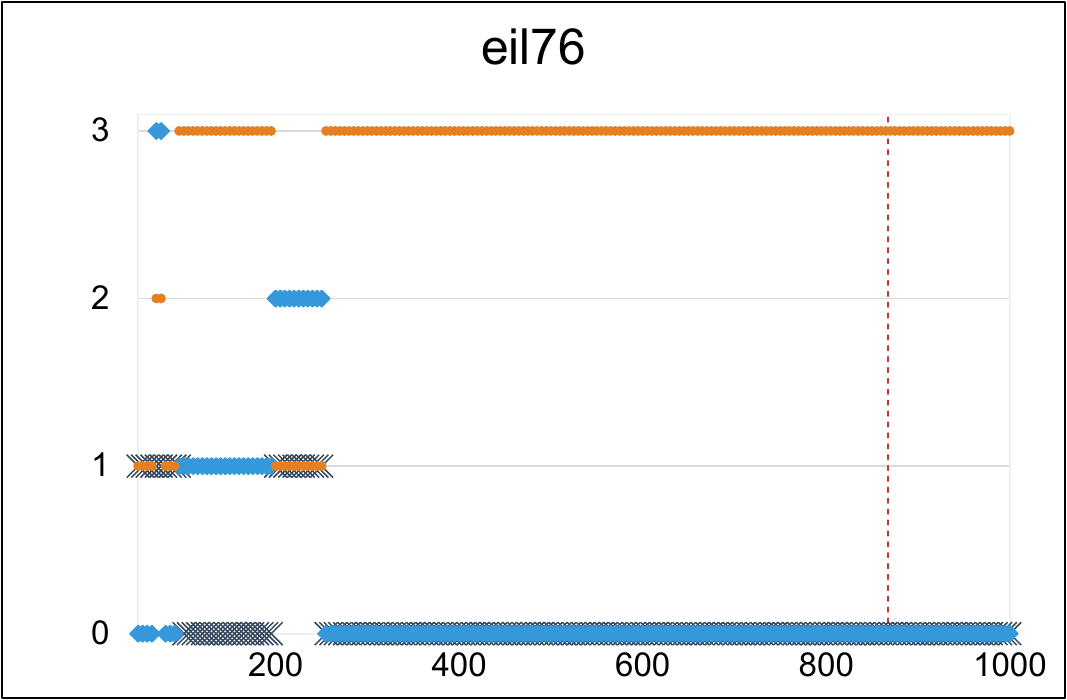}
    \end{subfigure}
    \hfill
    \begin{subfigure}{0.32\textwidth}
        \includegraphics[width=\textwidth]{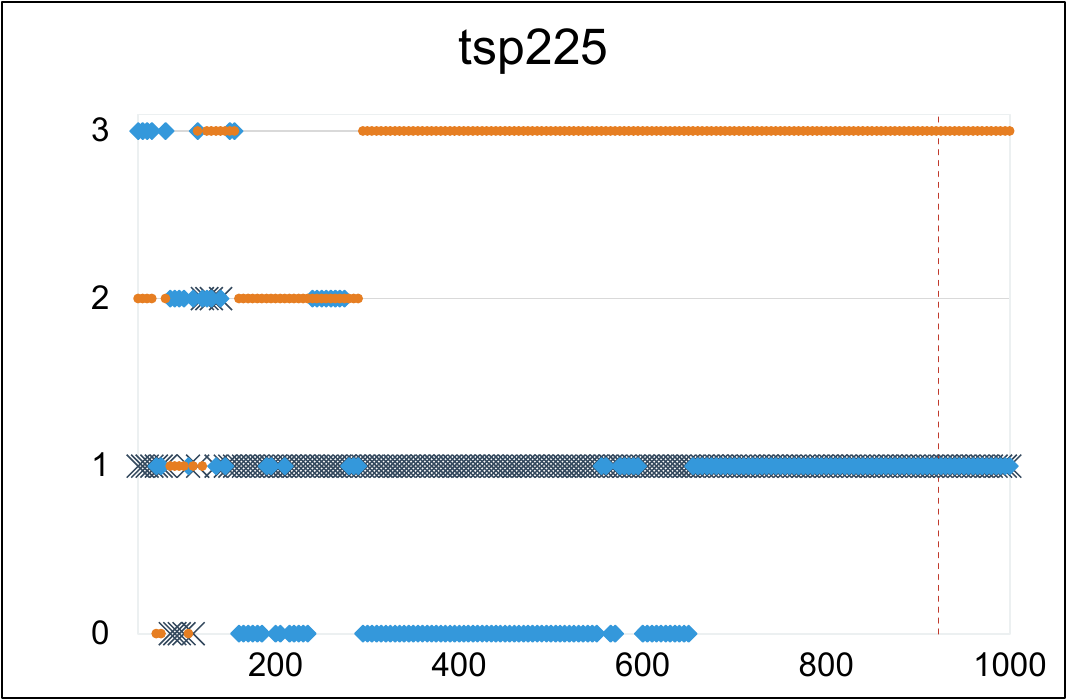}
    \end{subfigure}
    
    \caption{Rotation counts for {\color{flatlaci2} model-best (dark-blue crosses)}, {\color{flatblue2} iteration-best (blue diamonds)}, and {\color{flatorange} global-best (orange disks)} rotations selected by the best-performing MQEA-GH configurations at each time limit (ms). Vertical lines indicate the stabilization time limits.}
    \label{fig:rotationcount}
\end{figure*}

For synthetic instance classes, all three rotation types were used by the best-performing configurations for almost all tested time limits. In contrast, iteration-best rotations were frequently disabled for benchmark instances, while model-best rotations were also disabled over most time limits for eil76 and rd100. Global-best rotations generally received the highest count in the best-performing configurations for both synthetic and benchmark instances, often assuming the value $3$. When enabled, model-best rotations were generally assigned a count of one. For small time limits, larger model-best rotation counts were occasionally selected. However, beyond 200 ms, generally only a single model-best rotation was used. Iteration-best was the most context-sensitive rotation type. When it was not disabled, its count fluctuated depending on the instance and time limit. Figure~\ref{fig:rotationcount} presents representative graphs for the rotation-count selections. Notably, iteration-best rotations generally received higher counts than global-best rotations for synthetic classes with $\sigma=25$ under larger computational budgets, supporting the dependence of preferred rotation counts on instance characteristics.

\begin{figure*}[!t]
    \centering
    \begin{subfigure}{0.32\textwidth}
        \includegraphics[width=\textwidth]{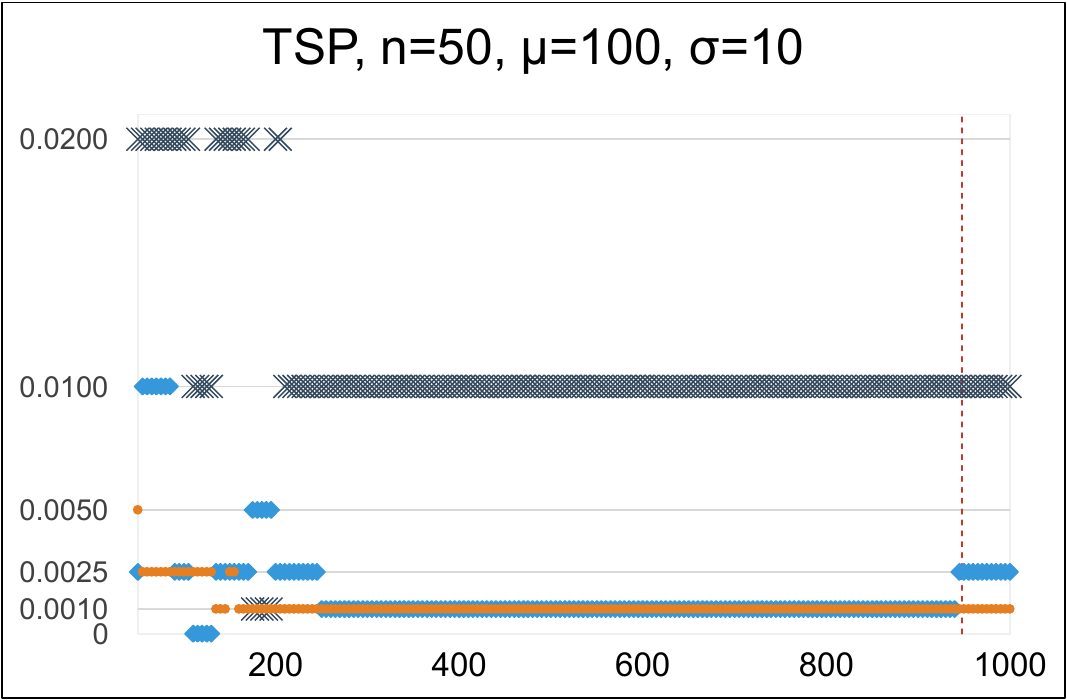}
    \end{subfigure}
    \hfill
    \begin{subfigure}{0.32\textwidth}
        \includegraphics[width=\textwidth]{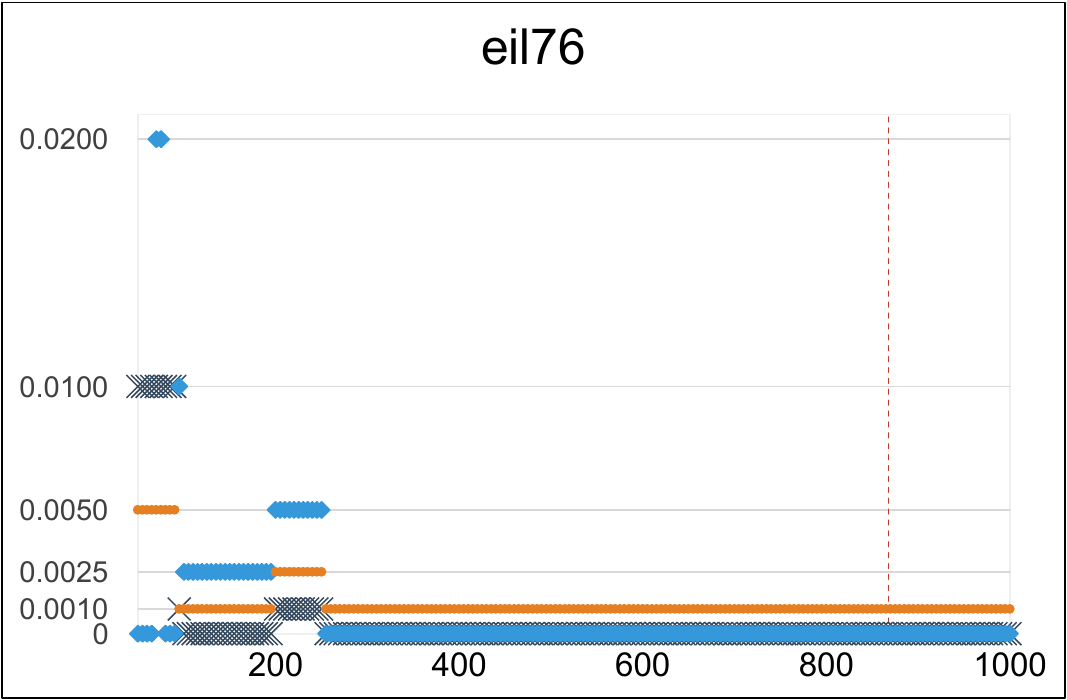}
    \end{subfigure}
    \hfill
    \begin{subfigure}{0.32\textwidth}
        \includegraphics[width=\textwidth]{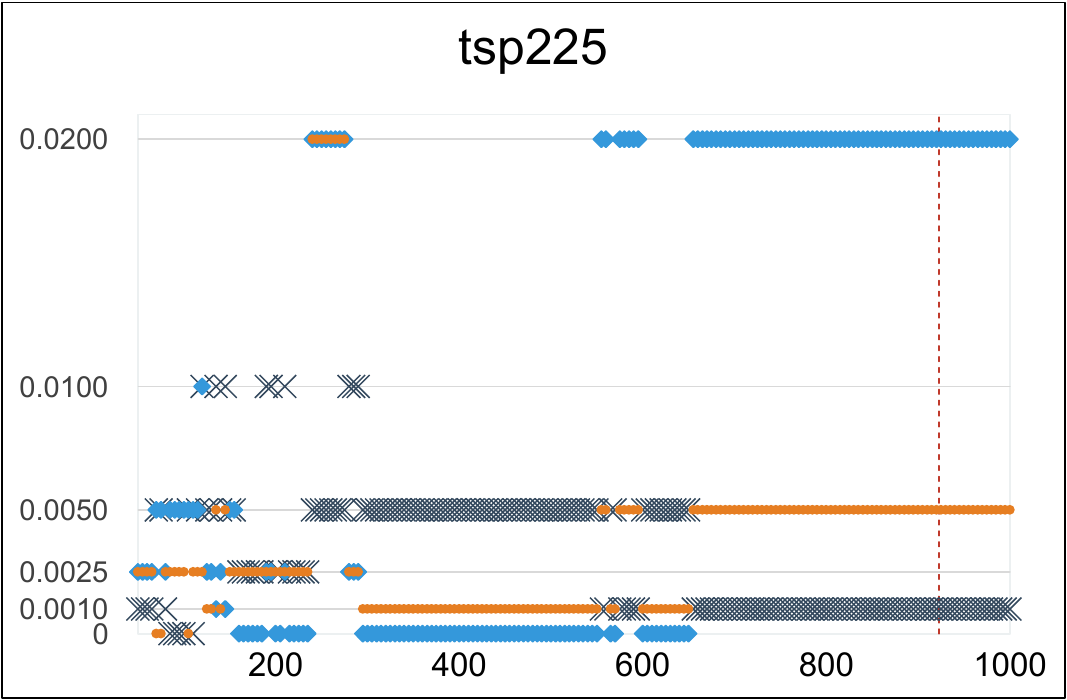}
    \end{subfigure}
    
    \caption{Base rotation angles for {\color{flatlaci2} model-best (dark-blue crosses)}, {\color{flatblue2} iteration-best (blue diamonds)}, and {\color{flatorange} global-best (orange disks)} rotations selected by the best-performing MQEA-GH configurations at each time limit (ms). Vertical lines indicate the stabilization time limits. Disabled rotation types are represented by a zero base rotation angle.}
    \label{fig:rotationangle}
\end{figure*}

\begin{figure*}[!t]
    \centering
    \begin{subfigure}{0.32\textwidth}
        \includegraphics[width=\textwidth]{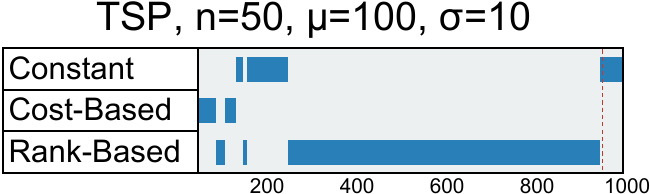}
    \end{subfigure}
    \hfill
    \begin{subfigure}{0.32\textwidth}
        \includegraphics[width=\textwidth]{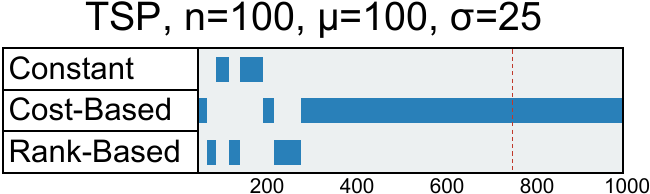}
    \end{subfigure}
    \hfill
    \begin{subfigure}{0.32\textwidth}
        \includegraphics[width=\textwidth]{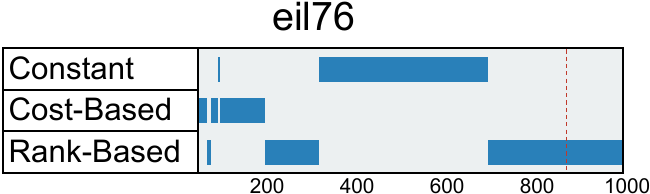}
    \end{subfigure}
    
    \caption{Selection of angle-calculation operators by the best-performing MQEA-GH configurations at each time limit (ms). Vertical lines indicate the stabilization time limits.}
    \label{fig:anglecalculation}
\end{figure*}

Smaller base rotation angles were generally preferred under larger computational budgets, which may reflect a preference for more gradual updates as the allowed computational budget increases. Global-best rotations generally received the lowest base angle values. Considered with the rotation counts, this indicates a preference for repeated global-best rotations, each with small base angles. In contrast, model-best rotations were generally assigned larger base angles, suggesting a preference for the use of fewer model-best rotations, each with relatively large base angles. When enabled, iteration-best rotations were generally assigned base angles greater than or equal to those of global-best rotations and less than or equal to those of model-best rotations. The main exception to both observations was tsp225, for which larger base angles were preferred for larger time limits, and iteration-best rotations received larger base angles than model-best under larger computational budgets. Figures~\ref{fig:rotationcount}~and~\ref{fig:rotationangle} illustrate these results.

Our final analysis reveals that the selection of angle-calculation operator depends strongly on instance characteristics and computational budgets. All three operators were utilized for substantial portions of time limits for various instances. However, for most instances, the operator selection did not vary after certain time limits, and a single operator was used for large portions of the time-limit range. Figure~\ref{fig:anglecalculation} presents these results.

\section{Conclusion and Future Work}
\label{sec:conclusion}

In this study, we proposed two mechanisms to improve the performance of multi-level quantum-inspired evolutionary algorithms for ordering problems. The first is a graph-based qudit model that captures dependencies between consecutive elements of a solution string without requiring entangled states. The second is heuristic initialization, which incorporates problem-specific information into the initial probabilistic representation to guide the search. We implemented a common algorithm framework for BQEA, MQEA, MQEA-G, and MQEA-GH, supporting a wide selection of configurations based on different operators and hyperparameters. We evaluated these variants on synthetic and benchmark instances of the TSP under various computational budgets. The results demonstrated the effectiveness of the proposed mechanisms, as MQEA-GH consistently outperformed the other QEA variants. Furthermore, MQEA-GH outperformed the established metaheuristics ACO, GA, and SA under most of the considered time limits on the synthetic instances, while generally showing competitive performance on benchmark instances. The experimental analysis also revealed that the preferred operators and hyperparameter values of MQEA-GH depend on instance characteristics and computational budgets, showing the importance of configuration selection in time-limited optimization.

A natural direction for future research is to apply
MQEA-GH and similar variants to other discrete optimization problems to examine the effectiveness of the
proposed methods in different settings. Another potential direction is to explore alternative probabilistic models that capture dependencies between the elements of candidate solutions, including models inspired by entangled states. Finally, developing hybrid variants of MQEA-GH by integrating mechanisms from other metaheuristics, such as local search, may offer further performance improvements.

\section*{Acknowledgments}

This work is funded by the German Federal Ministry of Research, Technology and Space within the funding program quantum technologies — from basic research to market — contract number 13N16090.


\bibliographystyle{plain}
\bibliography{references}
\end{document}